\documentclass[a4paper,fleqn]{cas-sc}
\usepackage{setspace}
\usepackage{flafter}

\usepackage[sort&compress,numbers]{natbib}

\usepackage[numbers]{natbib}
\usepackage{framed,multirow}

\usepackage{amssymb}
\usepackage{latexsym}
\usepackage{amsmath}
\usepackage{nccmath}
\usepackage{amsmath}
\usepackage[linesnumbered,ruled]{algorithm2e}
\usepackage{array}
\usepackage{graphicx}
\usepackage{multirow}
\usepackage{amsmath}
\usepackage{xcolor}
\newcommand*\rot{\rotatebox{90}}

\usepackage{url}
\usepackage{xcolor}
\usepackage{multirow}
\usepackage{array}
\usepackage{graphicx}
\def\tsc#1{\csdef{#1}{\textsc{\lowercase{#1}}\xspace}}
\tsc{WGM}
\tsc{QE}
\tsc{EP}
\tsc{PMS}
\tsc{BEC}
\tsc{DE}
\begin{document}
\let\WriteBookmarks\relax
\def\floatpagepagefraction{1}
\def\textpagefraction{.001}
\shorttitle{Efficient texture-aware multi-GAN for image inpainting}
\shortauthors{MA Hedjazi et~al.}

\title [mode = title]{Efficient texture-aware multi-GAN for image inpainting}                      

 \author[1]{Mohamed Abbas Hedjazi}[type=editor,
    auid=000,bioid=1,
    prefix=, 
    role=, 
    orcid=0000-0001-9492-3719]
  \cormark[1] 
  \ead{mahedjazi@gtu.edu.tr} 

  \author[1] {Yakup Genc}[type=editor,
    auid=000,bioid=1,
    prefix=, 
    role=, 
    orcid=0000-0002-6952-6735]
  \ead{yakup.genc@gtu.edu.tr} 

  \address[1]{Gebze Technical University, Kocaeli, 41400, Turkey}
  
  \cortext[cor1]{Corresponding author.} 

\begin{abstract}
Recent GAN-based (Generative adversarial networks) inpainting methods show remarkable improvements and generate plausible images using multi-stage networks or Contextual Attention Modules (CAM). However, these techniques increase the model complexity limiting their application in low-resource environments. Furthermore, they fail in generating high-resolution images with realistic texture details due to the GAN stability problem. Motivated by these observations, we propose a multi-GAN architecture improving both the performance and rendering efficiency. Our training schema optimizes the parameters of four progressive efficient generators and discriminators in an end-to-end manner. Filling in low-resolution images is less challenging for GANs due to the small dimensional space. Meanwhile, it guides higher resolution generators to learn the global structure consistency of the image. To constrain the inpainting task and ensure fine-grained textures, we adopt an LBP-based loss function to minimize the difference between the generated and the ground truth textures. We conduct our experiments on Places2 and CelebHQ datasets. Qualitative and quantitative results show that the proposed method not only performs favorably against state-of-the-art algorithms but also speeds up the inference time.

\end{abstract}



\begin{keywords}
Image inpainting  \sep Deep learning \sep Generative adversarial networks \sep Local binary pattern
\end{keywords}

\maketitle

\section{Introduction}
\label{introduction}

Image inpainting has attracted significant interest from computer vision and pattern recognition communities. It synthesizes plausible contents to fill in the missing regions or to remove unwanted objects from an image. It can be utilized in a wide range of applications, including image editing \cite{gated, visual2}, image compression \cite{compression}, image restoration \cite{restoration} and diminished reality \cite{dr}.

Infilling is a fundamental part of human vision. Vertebrate eyes do not cover the whole visual field due to a blind spot where optic nerves leave the eye. This spot does not contain any photo-receptor cells and does not contribute to the information flow of the scene. However, our brains use the information from the peripheral area, such as texture, geometry and semantics to fill the gap \cite{anstis2010visual}.

Prior approaches in computer vision solve the inpainting problem by extracting low-level features, matching and pasting patches \cite{PatchMatch, PatchMatch1, PatchMatch2}. These methods generate realistic textures in images with simple structures or small holes but usually present critical failures for images with non-repetitive patterns, such as faces and complex scenes.

Like many other tasks of computer vision, image inpainting also took its share with the rapid advancements in deep learning. Deep generative-based methods \cite{ce, globallocal, applied, visual} address the problems of traditional inpainting using generative adversarial networks (GANs) \cite{gan}. The latter demonstrates a powerful tool to fill in the corrupted image with plausible alternative contents by learning high-level features from large-scale datasets. However, most of the current GAN-based inpainting techniques suffer from problems related to structure preservation and unrealistic texture generation, which usually leads to blurry and geometrically distorted results. 

To address the aforementioned issues, most of the current GAN-based inpainting methods employ coarse\footnote{Coarse: initial prediction from the corrupted image. It contains fewer texture details.}-to-fine\footnote{Fine: generated by the refinement network that enhances the coarse image to have global consistency and fine-grained textures.} architectures \cite{gated, edge, structureflow, boosted}. Specifically, the coarse stage predicts the initial image from the corrupted one \cite{ca, MultiscaleGA} or reconstructs the image structure represented in the edge \cite{edge,structureflow}, the contour \cite{contour} and the segmentation labels \cite{semantic, boosted}. The refinement stage generally uses the predicted coarse image or the reconstructed information to generate a final plausible image. However, the performance of the mentioned multi-stage approaches is strongly related to the contour/edge/segmentation labels prediction stages. Furthermore, they require expensive computational resources since they optimize the parameters of two or more networks. Other studies employ the contextual attention mechanism (CAM) to borrow information from the surrounding parts \cite{ca, MultiscaleGA}. However, CAM still fails to ensure feature continuities \cite{semanticattention} and requires expensive computational resources.

In addition to the coarse-to-fine architecture and CAM, there exists another bottleneck that drastically increases the model capacity. Namely, training on high-resolution images, which involves big models with a large number of parameters. Consequently, the training becomes slower and enforces smaller batch sizes due to computational and memory resource constraints, which decreases the performances \cite{a}. Motivated by these observations, we introduce a new deep generative-based multi-resolution image inpainting framework to speed up the running time and restore fine-grained textures. Our approach is composed of four successive efficient generators filling in four different resolutions. Specifically, the training starts with lower-resolution images and progressively doubles their size, such that their corresponding generators can exploit the previously inpainted images (see Fig.~\ref{fig:architecture}). This improves the model stability since training GANs on low-resolution images proves easier and converges faster \cite{a}. Another main problem with direct high-resolution image synthesis is that the discriminator will focus on texture details. Hence, it can easily reject synthesized images in the early training stages.

Our approach drops the refinement module after the target resolution since it significantly increases the network size. We remedy the lack of this refinement stage by our proposed progressive training approach and a texture-based loss function. The latter adopts Local-binary-patterns (LBP) \cite{lbp} to the image inpainting task. LBP is a non-parametric texture descriptor that widely used in computer vision tasks \cite{lbpapp}. In particular, we minimize the distance between the ground truth LBP and the predicted one to enforce fine-grained textures in the corrupted regions. Hence, our approach does not require high computational resources since it neither performs complex operations (CAM) nor uses the refinement network. We conduct our qualitative and quantitative experiments on conventional inpainting datasets Places2 \cite{places} and CelebHQ \cite{style}. The results show that our efficient model can generate visually appealing images and outperforms current state-of-the-art methods. For this paper, the main contributions are as follows:

\begin{enumerate}
\itemsep=3pt
\item We present a new GAN-based image inpainting architecture that employs efficient progressive GANs to improve the performance and speed-up the inference time.
\item We adopt an LBP-based loss function to constrain the inpainting task and ensure realistic texture details.
\item The experiments on Places2 \cite{places} and CelebHQ \cite{style} datasets exhibit competitive qualitative and quantitative results against current state-of-the-art methods. We also show the scalability of the proposed approach to other applications, such as blind image inpainting and image outpainting.
\end{enumerate}

The remainder of this paper is organized as follows: Section~\ref{related_works} reviews the related work, section~\ref{preliminaries} presents the preliminaries of the study. Section~\ref{approach} explains our approach in detail. Section~\ref{experimental} describes the experimental evaluation and provides quantitative and qualitative comparisons against the state-of-the-art methods. Finally, Section~\ref{conclusion} presents the conclusions and directions for future work.

\section{Related works}
\label{related_works}
Many image inpainting approaches are proposed in the literature. They can be classified into two major categories: traditional and deep learning approaches. Traditional methods employ either diffusion-based or patch-based techniques. Diffusion-based techniques fill in the holes by propagating the appearance of the neighborhood region to them \cite{isophotes}. Therefore, they may fail to generate meaningful structures for large or complex holes since only surrounding pixels of missing parts contribute to the inpainting process. On the other hand, patch-based image inpainting can fill in relatively larger corrupted regions with realistic textures by searching and copying the best matching patches \cite{PatchMatch}. However, this iterative operation is expensive in terms of both memory and time. To overcome this limitation, \cite{PatchMatch1} generalizes the previous algorithm and speeds up the inpainting application. Furthermore, patch-based methods extract only low-level features. Consequently, they can not understand the semantic structure of the image resulting in low performances in many cases, such as images of crowded scenes. Another approach \cite{multimedia} proposes a robust forgery detection algorithm of image inpainting. It combines the joint probability density matrix with a former object removal approach \cite{liang} to resist postprocessing attacks such as Gaussian noise and JPEG compression.

Learning-based methods benefit from the fast improvements of deep neural networks (DNNs) and GANs \cite{gan} to learn the image semantic from large-scale datasets \cite{pyramid, progressive_rec, recurrent}. These methods directly predict the missing pixel values using encoder-decoder architectures. Context encoders (CE) \cite{ce} is one of the first attempts that fill in a square hole in the center of the image using adversarial learning. The method suffers from significant artifacts and exhibits blurriness. CE was improved by \cite{globallocal} using two discriminators to ensure global and local image consistency. A postprocessing step using \cite{telea2004image} followed by \cite{perez2003poisson} is required to guarantee the color coherency around square holes. \cite{ca} replaces the postprocessing step by attaching the coarse network to another refinement network, which employs the contextual attention mechanism (CAM). This method enhances the semantic consistency since it searches for a collection of surrounding background patches with the highest similarity score to the coarse image. However, it does not ensure pixel continuities since it is trained using rectangular regions. This was addressed by \cite{semanticattention} that can handle free form masks by adding a coherent semantic attention layer to the refinement network. However, this method is time-consuming since it performs complex operations requiring high computational resources. \cite{visual} reduces the number of the parameters using a squeeze-and-excitation residual network in both generator and discriminator. Besides, it proposes a joint context-awareness loss to generate more realistic textures. Other approaches handle irregular masks and address the artifacts problem without using adversarial learning. \cite{partial} employs an automatic mask updating mechanism of the partial convolution layers that eliminate substituting pixels and use only valid pixels. \cite{dfnet} achieves competitive results using a fusion block that generates a flexible alpha composition map to combine corrupted and non-corrupted pixels. Also, it uses UNet architecture embedded with the proposed fusion blocks to handle nonharmonic region boundaries. \cite{applied} employs global and local discriminators to build a fusion network that produces semantically coherent images. 

Other methods utilize multi-stage architectures to reduce the complexity of the inpainting problem by providing additional information to the model. \cite{semantic} is a two-stage architecture that predicts the segmentation labels to generate plausible images of foreground objects. \cite{contour} is a three-stage architecture that uses the contour information to preserve both foreground and background object boundaries. In another two-stage architecture \cite{edge}, the edges are predicted to supervise the model prediction and recover the image structure. \cite{structureflow} adds appearance flow to a second stage to establish long-term corrections between masked and contextual regions. The hand-drawn sketches and gated layers generate plausible images using free from masks in \cite{gated}. \cite{MultiscaleGA} is also a coarse-to-fine architecture that predicts a high-resolution coarse image in the first stage. Later, it uses a refinement generator with multi-scale discriminators to generate smooth images. However, the use of the attention layer in the refinement network significantly increases the computational complexity that augments the inference time of \cite{MultiscaleGA, gated}. Reducing the model size without affecting the quality of the generated images is desirable. We present a GAN-based inpainting method that reduces the inference time and generate plausible results. It uses neither CAM nor coarse-to-fine architectures.

\begin{figure}
\centering
\includegraphics[width=0.6\textwidth]{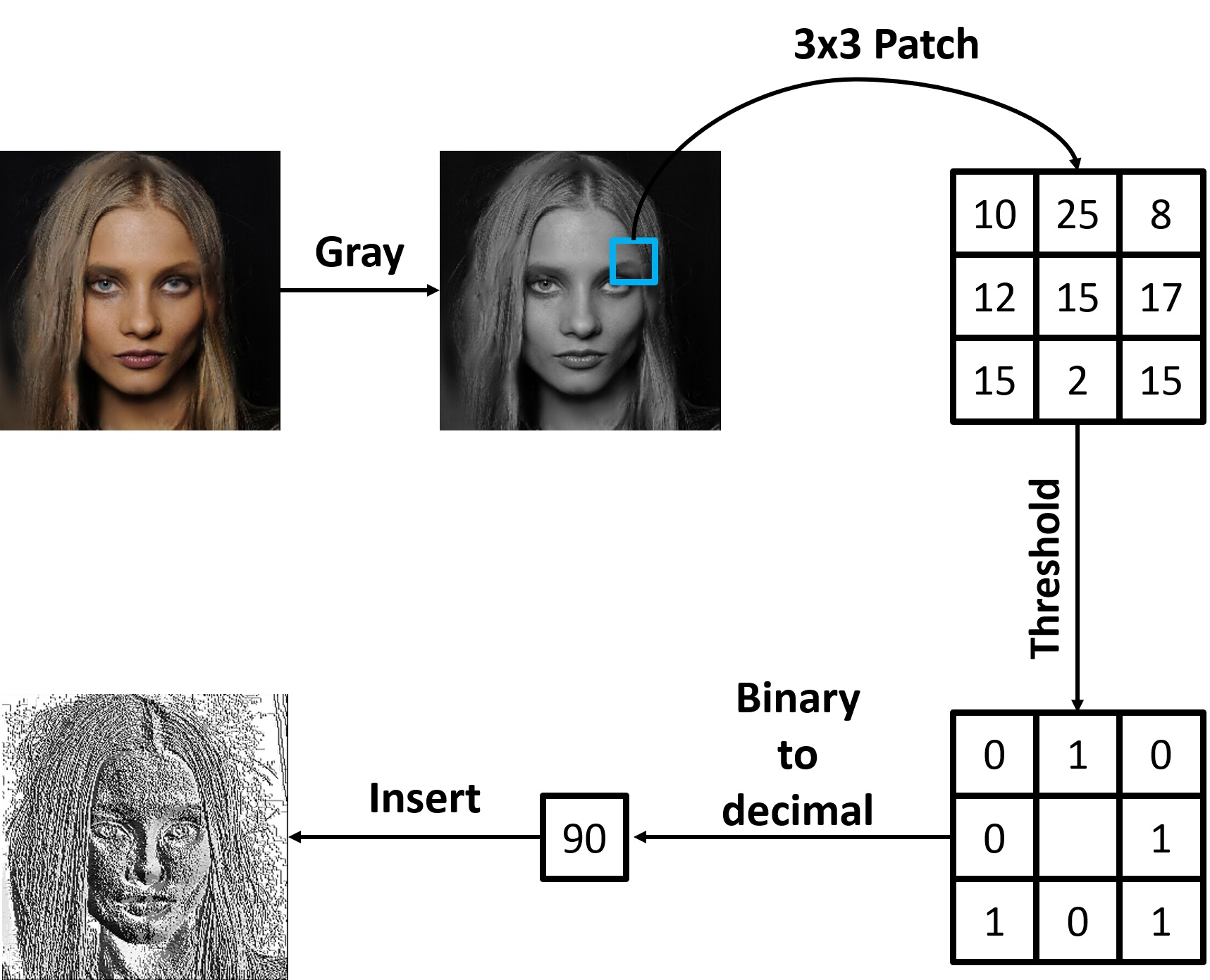}
\caption{Example of a $3\times3$ LBP operator applied on a gray-scale image (This figure should be printed in color).}
\label{fig:lbp}
\end{figure}

\section{Preliminaries}
\label{preliminaries}
We adopt GANs \cite{gan} and the LBP operator \cite{lbp} to design a new end-end-image inpainting framework. Specifically, we present an efficient progressive GANs architecture to stabilize the training, improve the performance and speed-up the inference time. Furthermore, we adopt an LBP-based loss function to constrain the inpainting task and ensure fine-grained textures.

\subsection{Generative adversarial networks}
\label{gan}
Introduced in \cite{gan}, GANs have shown huge success in image synthesis and have been adopted for modeling complex computer vision problems, including video generation \cite{videogeneration}, image-to-image translation \cite{pix2pix} and modulation classification \cite{semi}. Although GANs are rapidly improving and building a new state-of-the-art in these tasks, they are still hard to train. They optimize the parameters of two neural networks independently in a mini-max game. The first network is a generator that produces new samples similar to the real data. The discriminator network is optimized to distinguish between fake and real data. The loss function is defined as follows:

\begin{equation} 
\label{eq:dis_nsgan}
\min _{G}\max _{D}E_{x\sim P_{data}(x)}[\log \left(x\right)] + E_{z\sim P_{z}(z)}[\log \left( 1-D(G(z\right))]
\end{equation}

Where $z$ can be a vector sampled from a Gaussian distribution in random image generation tasks or an image in image-to-image translation tasks, $x$ is a real data sample, $G(.)$ is the generator network, and $D(.)$ is the discriminator network.

\subsection{Local binary patterns}
\label{lbp}
LBP is a non-parametric image operator that transforms an image into an array representing the local structure of the image by comparing each pixel with its adjacent pixels \cite{lbp}. LBP is a robust descriptor that can summarize the most important texture information in an image. Also, it shows computational simplicity and good performance in many computer vision and image processing applications \cite{lbpapp}. An example of a $3\times3$ LBP operator is shown in Fig.~\ref{fig:lbp}. LBP iterates over each pixel in a gray-scale image to check the values of the surrounding $3\times3$ patch, whether they are smaller than the center pixel or not. The resulting binary number is converted to a decimal number and placed in the corresponding position in the LBP image.

\begin{figure}
\centering
\includegraphics[width=0.99\textwidth]{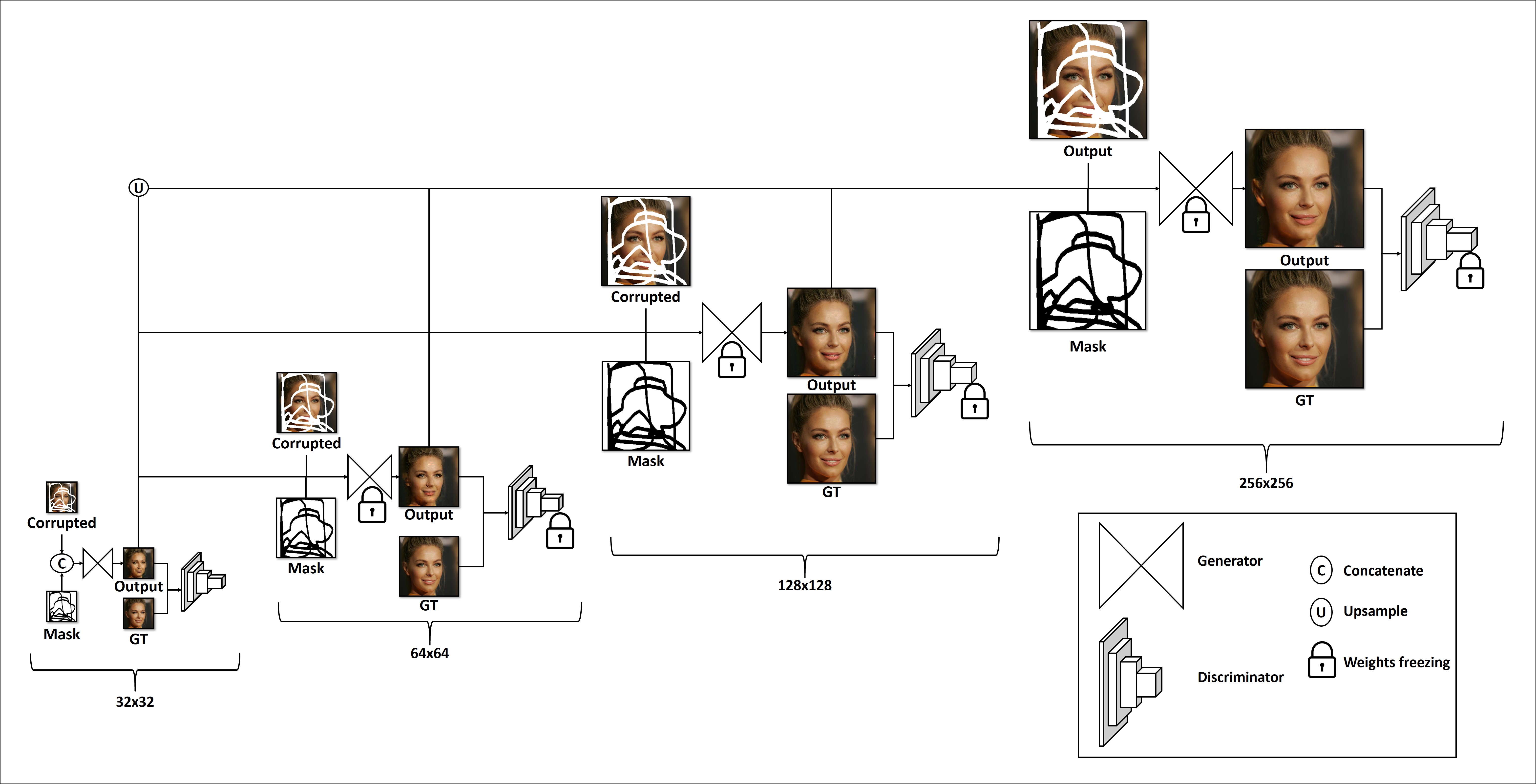}
\caption{Overview of our proposed network architecture. It has four progressive generators (32, 64, 128 and 256), such that the generator of each scale exploits the previously inpainted image resolutions to fill in the corrupted image. We use four PatchGAN \cite{pix2pix} discriminators to learn how to tell apart real from generated images. When training the current resolution network, we freeze the weights of all the generators/discriminators of the other resolutions.}
\label{fig:architecture}
\end{figure}

\section{Approach}
\label{approach}
\subsection{Multi-resolution-based inpainting}
\label{approach:multi-resolution}
Training GANs on high-resolution images is a challenging optimization problem that involves millions of parameters. \cite{a} produces low-resolution images from a latent Gaussian vector in the first stage. During training, it progressively adds layers to the generator and the discriminator to increase the image resolution. However, this framework is not suitable for image-to-image translation applications since they require a high-resolution image as input. We introduce a GAN-based architecture for image inpainting that includes four progressive generators and discriminators. We train an encoder-decoder generator on a low-resolution image for many epochs to robustly produce samples with a very close distribution to the original one. As the training progresses, we use the pretrained generators as the starting point for the successive higher-resolution generator. Using this strategy helps the latter to exploit the filled-in regions of the previous lower-resolution images to learn the global image consistency and complete the corrupted regions with correct structures. In contrast, training GANs on high-resolution images is hard to stabilize, which may affect the performance of the model. We can explain this by the fact that, during training, the discriminator keeps rejecting most of the generated images, since the ground truth image contains fine-grained texture details which are very difficult for the generator to produce \cite{a, odena2017conditional}. 

To the best of our knowledge, the proposed architecture is one of the first studies that apply progressive generators and discriminators for image inpainting. \cite{MultiscaleGA} is a coarse-to-fine architecture that predicts a high-resolution coarse image and enhances it using multi-scale discriminators in the refinement stage. The discriminator of each scale criticizes the output of that particular resolution size and gives high gradient feedback to early convolution layers. However, it does not directly exploit the refined images in lower scales, which may still be a bottleneck for the high-resolution discriminator to easily reject the generated samples. In contrast, we build our high-resolution prediction on already filled-in predictions in lower-resolutions (for more details see section \ref{approach:architecture} and Fig.~\ref{fig:architecture}). In this way, the discriminator criticizes reasonable synthesizes images that are close to real samples. The proposed approach neither uses coarse-to-fine architecture nor an attention mechanism that significantly increases the model complexity in \cite{MultiscaleGA}. Another approach in \cite{dfnet} uses a UNet architecture embedded with fusion blocks in a multi-scale manner. However, they drop the adversarial learning and use perceptual and style losses \cite{perceptual} to enforce texture details. Using only the former losses without adversarial learning can result in checkerboard artifacts since it is hard to find the best loss weights \cite{partial}. In our approach, we use the adversarial, the reconstruction and the proposed LBP loss functions to enhance the image texture (see section \ref{approach:lbp}). 

\begin{figure}
\centering
\includegraphics[width=0.6\textwidth]{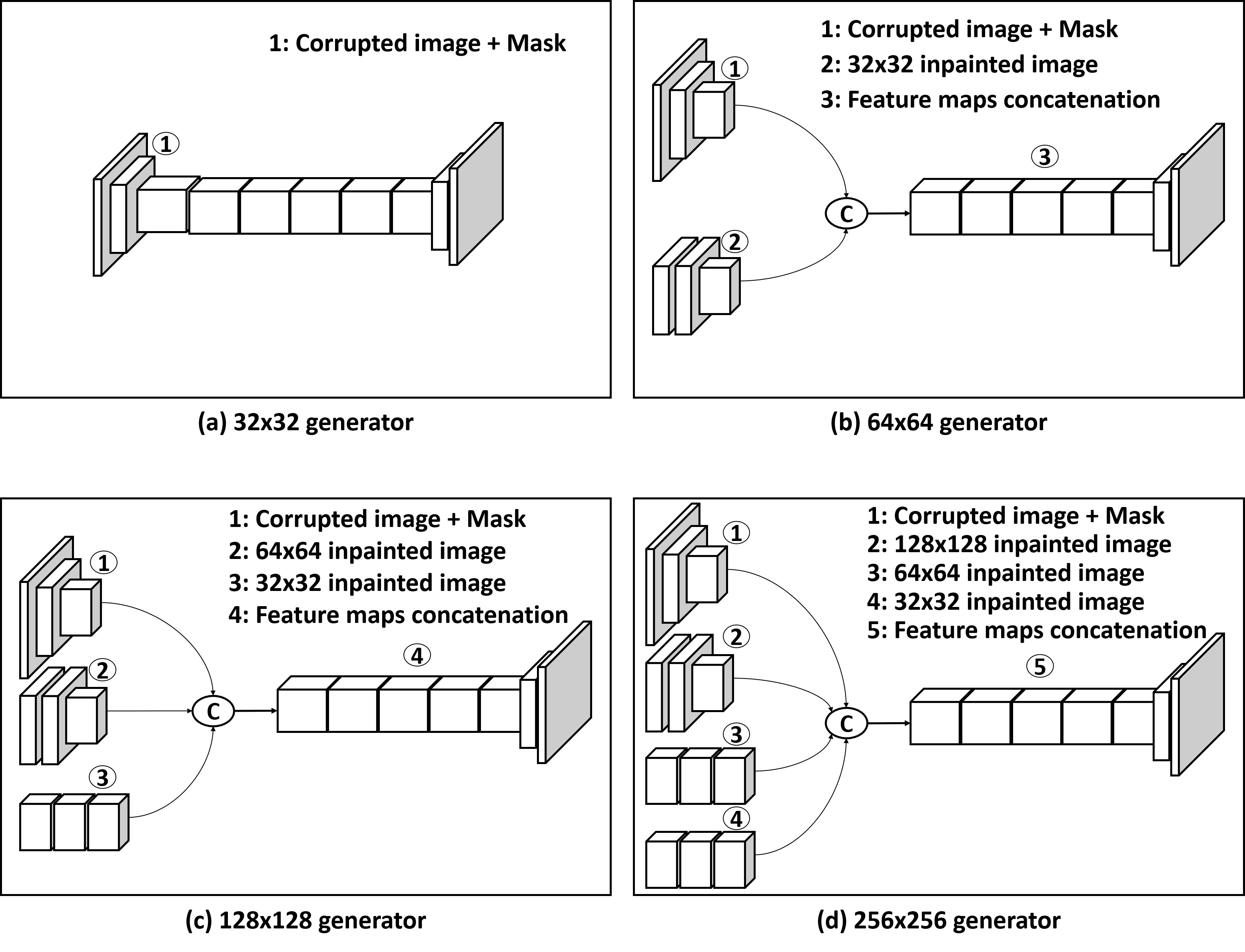}
\caption{Sub-figures a, b, c and d illustrate the input and the output of the generators of the four different resolutions 32, 64, 128, and 256.}
\label{fig:generators}
\end{figure}

\subsection{Architecture}
\label{approach:architecture}

Using different resolutions investigates multiple receptive fields, which helps the network to learn the global structure of the image. As described in Fig.~\ref{fig:architecture}, the training starts with the $32\times32$ resolution images. We channel-wise concatenate the corrupted image and the mask to feed them to their specific-resolution generator. The output and the ground truth images are then given to the PatchGAN \cite{pix2pix} discriminator, which shifts the generator distribution to the real one. We train the generator network defined in Fig.~\ref{fig:generators} (a) until convergence, and we visualize different quantitative metrics and loss values. We stop the training when the metrics become stable, and the produced images are visually good enough to be used as input for the next resolution. We use the pretrained generator of the $32\times32$ resolution to train the next resolution network ($64\times64$). The network in Fig.~\ref{fig:generators} (b) contains three sub-networks. We feed the concatenation of the corrupted image and the mask to the first sub-network. We feed the predicted image of resolution $32\times32$ to the second sub-network. The last sub-network takes the concatenation of the feature maps produces by the previous sub-networks to produce the final image ($64\times64$). Similarly, we follow the same approach for the last two resolutions ($128\times128$ and $256\times256$), where each generator exploits the previously inpainted images as described in Fig.~\ref{fig:generators} (c) and Fig.~\ref{fig:generators} (d). Using the proposed architecture in the encoder/decoder networks (see Appendix.~\ref{appendix_generators}) reduces the number of parameters while improving the performance. This simple architecture for encoders is not a good candidate to benefit from depth or width scaling ideas in EfficientNet \cite{efficientnet} since the latter is designed for huge classification models.

\subsection{LBP for texture preservation}
\label{approach:lbp}

Early deep-based inpainting methods assume that the image texture and semantic can be learned automatically by CNNs without further supervision. Recent GAN-based methods demonstrate that this task is challenging and require additional information. \cite{b} employs discriminative modules and class supervision to enforce fine-grained features. Other GAN-based inpainting approaches add \cite{gated} or predict \cite{edge,structureflow} edges to ensure realistic textures. However, choosing the right threshold for the Canny edge detector \cite{canny} that can preserve the image texture for both highly structured and simple images is difficult in practice. Furthermore, the edges can not provide sufficient texture details in many cases, such as the face skin and uniform backgrounds.

Motivated by these observations, we investigate hand-crafted features. Specifically, we adopt the famous texture operator LBP \cite{lbp} as a new loss function for image inpainting to ensure better texture learning as used in \cite{LBP-BEGAN} for infrared and visible image fusion and \cite{WeightOptimalLBP} for face recognition. Particularly, we minimize the loss between the LBP of the ground truth and the predicted images using the LBP layer defined in Algorithm~\ref{alg:lbp}. We select the LBP operator since it is robust to illumination variations and invariance to gray-scale changes. Furthermore, it does not add parameters to the network and it is computationally nonexpensive. However, LBP is a non-differentiable iterative function that can not be optimized using backpropagation. To address that, we transform the problem into matrix multiplication operations using a fixed weight convolution layer. Thus, it does not add learnable parameters to our full model. We base our implementation on \cite{lbpdiff}. Note that we only use the LBP loss in the last resolution ($256\times256$), which speeds up the inference time.

\begin{algorithm}
\underline{class LBPLayer}\;
\underline{function initialize}$()$\;
- Conv: 2D convolution layer, where $in_{channels}=1$, $out_{channels}=8$, $kernel=3$, $stride=1$, $dilation=1$, $bias=False$\;
- Initialize the kernels to zeros\;
- Initialize the center of the kernels to -1\;
- Initialize the remaining values of each kernel to 1\;
- Codes: list of 8 values initialized from $2^0$  to  $2^7$\;
- ReLU: Rectified Linear Unit activation function\;

\underline{function forward}$(image_{gray})$\;
\SetKwInOut{Input}{Input}
\SetKwInOut{Output}{Output}
\Input{$image_{gray}$}
\Output{$image_{lbp}$}
- $result$ = $Conv(image_{gray})$.\;
- $result$ = $ReLU(result)$.\;
- $result$ = $result * Codes$.\;
- $result$ = $result.sum(dim=1)$.\;
- $image_{lbp}$ = $result / 255$.\;
\KwRet $image_{lbp}$\;
\caption{The LBP layer pseudo-code}
\label{alg:lbp}
\end{algorithm}

\subsection{Loss functions}
\label{loss_functions}
Let $I_{n\times n}$ and $M_{n\times n}$ be the ground truth image and the mask, where $n$ is the size of a square image. Also, let $G_{n\times n}(.)$ be a generator network that produces an image $O_{n\times n}$. Let $Gray(.)$ be a function that transforms a color image into a gray-scale image. Let $LBP(.)$ be a differentiable LBP layer that takes a gray-scale image and outputs the LBP image. The output image for various resolutions can be obtained using Equations ~\ref{eq:gen_32x32}-\ref{eq:gen_256x256}.

\begin{equation}
\label{eq:gen_32x32}
O_{32\times32} =G_{32\times32}(I_{32\times32} * M_{32\times32}, M_{32\times32})
\end{equation}

\begin{equation}
\label{eq:gen_64x64}
O_{64\times64} =G_{64\times64}(I_{64\times64} * M_{64\times64}, M_{64\times64}, O_{32\times32})
\end{equation}

\begin{equation}
\label{eq:gen_128x128}
O_{128\times128} =G_{128\times128}(I_{128\times128} * M_{128\times128}, M_{128\times128}, O_{32\times32}, O_{64\times64})
\end{equation}

\begin{equation}
\label{eq:gen_256x256}
O_{256\times256} =G_{256\times256}(I_{256\times256} * M_{256\times256}, M_{256\times256}, O_{32\times32}, O_{64\times64}, O_{128\times128})
\end{equation}

\textbf{L1 loss:} we measure the error between the ground truth image and the predicted image for each resolution as defined in Eq.~\ref{eq:rec}.

\begin{equation}
\label{eq:rec}
L_{rec} = ||O_{n\times n} - I_{n\times n}||_1
\end{equation}

\textbf{Adversarial loss:} we optimize the LSGAN \cite{lsgan} adversarial loss for each resolution as defined in Eq.~\ref{eq:dis} and Eq.~\ref{eq:gen}, respectively.

\begin{equation} 
\label{eq:dis}
L_{dis}=E[(D(I_{n\times n}) - 1)^2] + E[D(O_{n\times n})^2]
\end{equation}

\begin{equation} 
\label{eq:gen}
L_{adv}=E[(D(O_{n\times n}) - 1)^2]
\end{equation}

\textbf{Texture loss:} we use the LBP differentiable layer to calculate the loss between the ground truth texture and the generated $256\times256$ image texture, see Eq.~\ref{eq:texture}.

\begin{equation} 
\label{eq:texture}
L_{texture} = ||LBP(Gray(O_{fine})) - LBP(Gray(I_g))||_1
\end{equation}

\textbf{Overal loss:} we use a weighted sum of the reconstruction, the adversarial and the texture loss. We give a weight of $\lambda_{adv}=0.1$, $\lambda_{rec}=1$ and $\lambda_{texture}=10$ for the adversarial loss, the reconstruction loss and the texture loss, respectively. The overall loss is defined in Eq.~\ref{eq:overall}.

\begin{equation} 
\label{eq:overall}
L_{overall} = \lambda_{adv} * L_{adv} + \lambda_{rec} * L_{rec} + \lambda_{texture} * L_{texture}
\end{equation}

\section{Experiments}
\label{experimental}
\subsection{Datasets and image masks}
\label{datasets}
We conduct our experiments using two conventional image inpainting datasets. The first one is Places2 \cite{places} that has more than 1.8M images and 400 scene categories, such as bedrooms, streets, etc. Although the Places2 dataset was created for the image classification task, it became a popular image inpainting dataset since it has a vast natural scene variation. We use the original train and test split for the Places2 dataset. To further enrich our experiments, we evaluate our method on CelebHQ \cite{style}, which is a challenging dataset that has 30K cropped face images selected from the CelebA \cite{celeba}. It has a large pose and background variations. We use the same training and test split of the CelebA dataset. Since users of image inpainting applications usually want to edit or remove arbitrary shapes in the scenes, we use irregular mask sizes \cite{Iskakov} during training. In test time, we classify the mask images based on the ratio between the hole size and the image size into four categories (10-20\%, 20-30\%, 30-40\%, and 40-50\%).

\subsection{Implementation details}
\label{implementation}
In this part, we describe our training procedure and the hyper-parameter settings. We use Pytorch v1.6 \cite{pytorch} to implement the proposed method using CUDA v10.1 and cuDNN v7.6.4. We use the Adam optimizer \cite{adam} with hyper-parameters $\alpha=0.5$ and $\beta=0.99$, respectively. We set the batch size to 32, and we fix the learning rates to $10^{-4}$ for the generators and the discriminators. We use spectral normalization \cite{spectral} in all the convolution layers of the discriminator. The details of the architectures illustrated in Fig.~\ref{fig:architecture} are described in Appendix~\ref{appendix_discriminators} and Appendix~\ref{appendix_generators} for the discriminators and the generators, respectively. We freeze the weights of the previous networks when training the generator and the discriminator of the current resolution.

\begin{figure}
\centering
\includegraphics[width=0.99\textwidth]{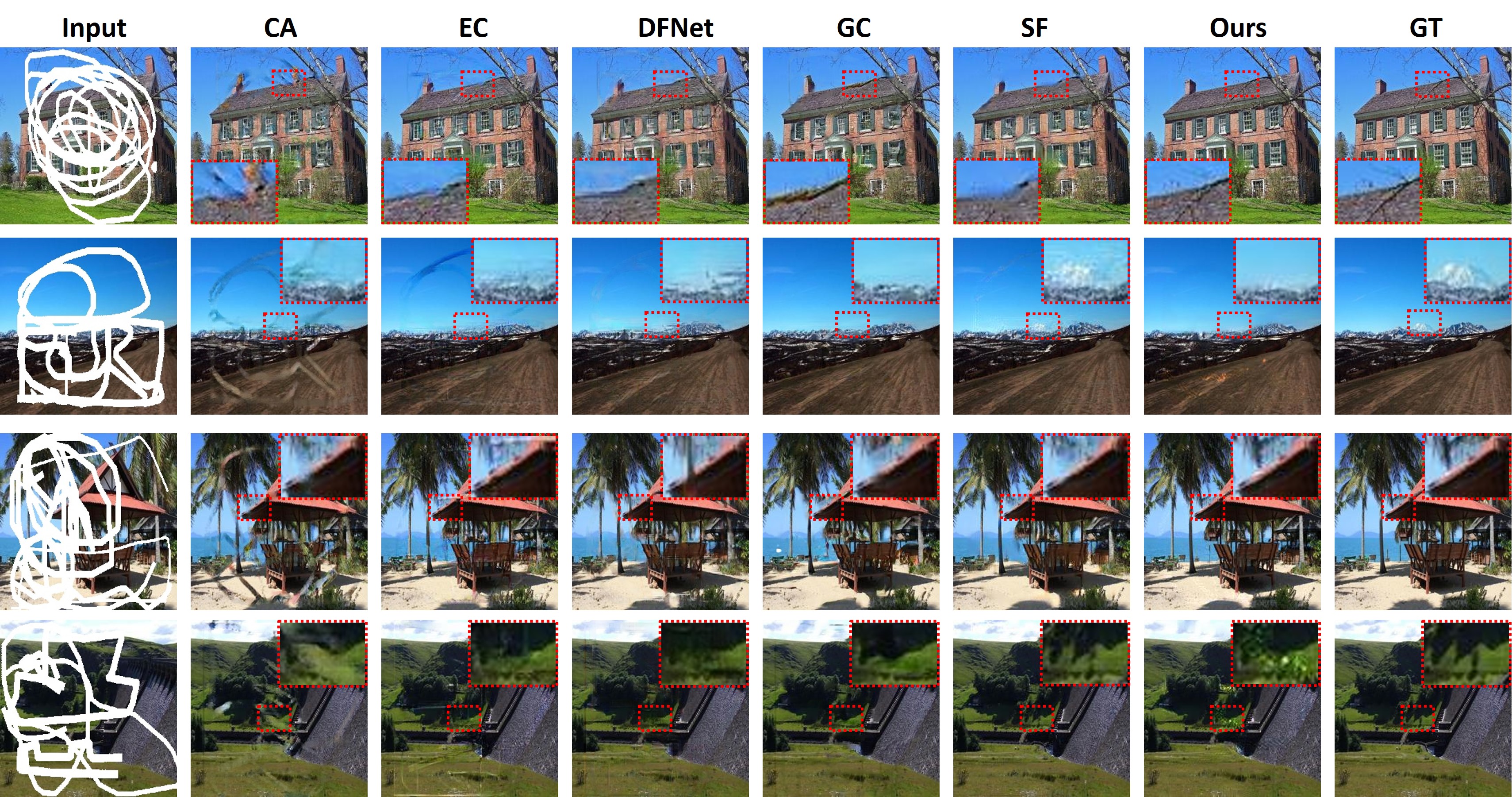}
\caption{Example cases of qualitative comparison between our model and state-of-the-art methods using irregular hole inpainting on Places2 \cite{places} test set. From left to right, we show the corrupted image and the results of CA \cite{ca}, EC \cite{edge}, DFNet \cite{dfnet}, GC \cite{gated}, SF \cite{structureflow}, our model and the ground truth image (This figure should be printed in color).}
\label{fig:places2}
\end{figure}

\begin{figure}
\centering
\includegraphics[width=0.99\textwidth]{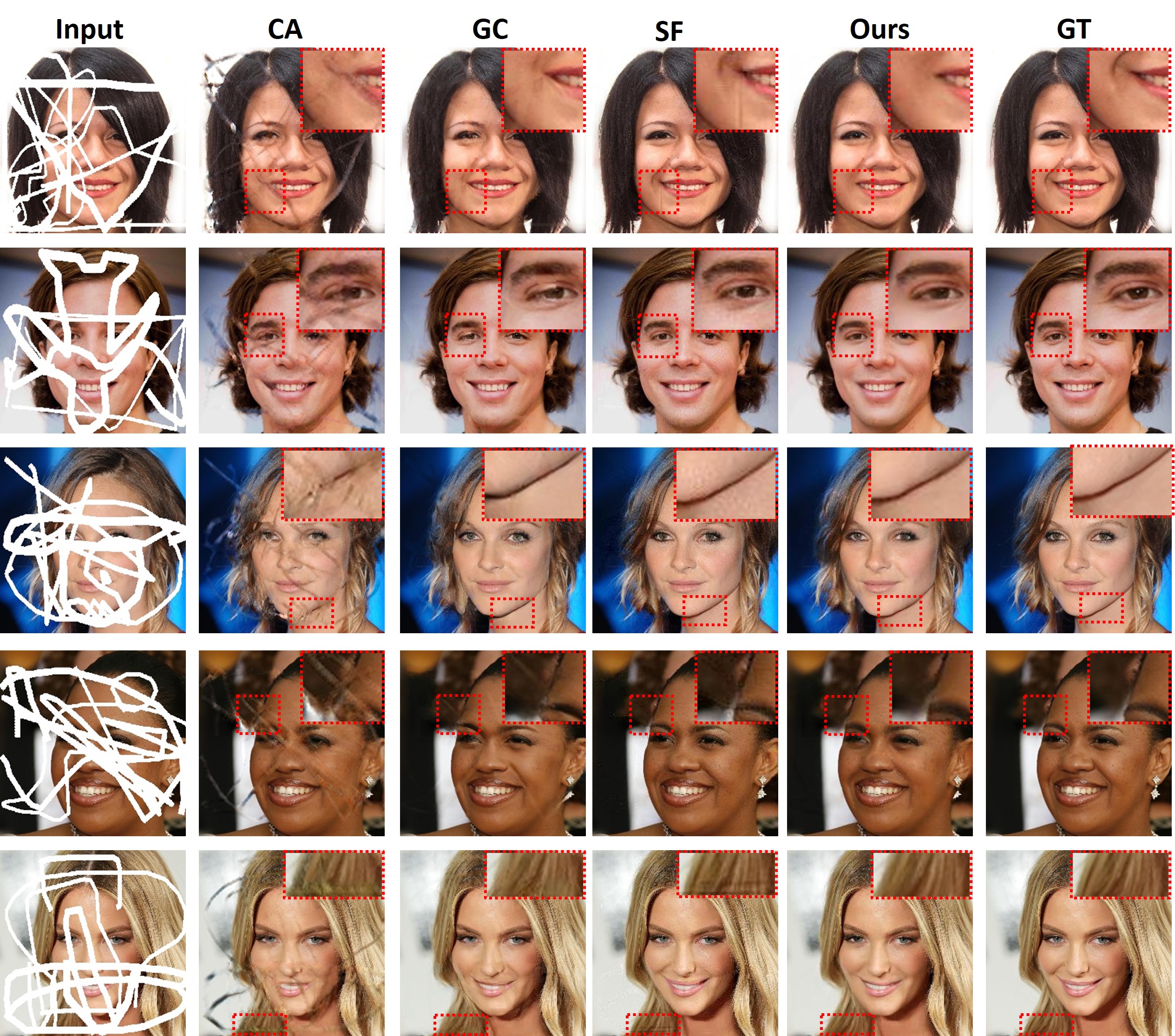}
\caption{Example cases of qualitative comparison between our model and state-of-the-art methods using irregular hole inpainting on CelebHQ \cite{style} test set. From left to right, we show the corrupted image, the results of CA \cite{ca}, GC \cite{gated}, SF \cite{structureflow}, our model and the ground truth image (This figure should be printed in color).}
\label{fig:celebhq}
\end{figure}

We qualitatively and quantitively compare our full model with current state-of-the-art methods, including Contextual Attention (CA) \cite{ca}, Edge Connect (EC) \cite{edge}, Deep Fusion Network (DFNet) \cite{dfnet}, Gated Convolution (GC) \cite{gated} and Structure Flow (SF) \cite{structureflow}. We select these approaches for two main reasons. The first one is that they have the pretrained models, which ensure a fair comparison and save both time and computational resources. The second reason is that they achieve very competitive results using different approaches. We use the original train and test splits for Places2 \cite{places} and CelebHQ \cite{style} datasets.

\subsection{Qualitative comparison}
\label{qualitative}
We qualitatively compare our approach with the selected state-of-the-art methods on two datasets. We zoom in different parts of the image to show the difference between the generated images. Seen from figure Fig.~\ref{fig:places2}, CA \cite{ca} generates significant artifacts leading to misrepresented structures. EC \cite{edge} produces better results since it predicts edges to recover the global structure of the image, but obvious visual artifacts still appear in the masked regions. While DFNet \cite{dfnet} generates plausible and smooth images with global image consistency using their introduced fusion blocks, it still exhibits observable color discrepancies. GC \cite{gated} produces realistic images due to the gated convolution layers and the refinement network, but it can miss some texture details. SF \cite{structureflow} generates plausible images with fine-grained textures since it employs two effective stages to preserve both structures and textures, respectively. However, it suffers from remarkable inconsistencies near the boundaries. Our method presents competitive results and shows very realistic textures in all the missing regions.

To further demonstrate the effectiveness of the proposed method, we report qualitative results on the CelebHQ dataset. We can see from Fig.~\ref{fig:celebhq} that the images produced by CA show visually poor performance. GC generates realistic images but still shows discordance between the background and the parts of the hole. SF \cite{structureflow} synthesis smooth faces with realistic textures. However, it sometimes exhibits color and row discontinuities in the predicted pixels. Our method presents the most natural faces without using large models or complex mechanisms such as CAM. The results can be explained in that our approach looks at different image scales using multiple GANs to ensure visually realistic images with both local and global structure consistencies. Meanwhile, the proposed LBP-based loss function both improves and sharpens the texture of the generated parts. Additional results of our proposed approach are present in the appendix (see Fig.~\ref{fig:additional_celebhq} and Fig.~\ref{fig:additional_places2}) showing that our model may synthesize plausible new contents due to adversarial learning. Furthermore, our qualitative results show that our model enforces close LBP features if the generated contents have resembling structures to the ground truth or only the color is changed. In contrast, the generated LBP features are distinct for modified structures. Note that in both cases, the LBP-based loss function ensures fine-grained textures.

\subsection{Quantitative comparison}
\label{quantitative}
To quantify the performance of the proposed approach, we use three well-known assessment metrics, including mean absolute error (MAE), peak signal-to-noise ratio (PSNR), and structural similarity index (SSIM) following works of \cite {edge, structureflow}. To achieve a fair comparison, we use the same masks and test splits of the two datasets. Table~\ref{places2} lists the evaluation results on the Places2 dataset. We can see that CA \cite{ca} shows the worst performances in the three metrics on different mask sizes. EC \cite{edge} exhibits the best results since it predicts the edges to supervise the image structure generation. The scores of DFNet \cite{dfnet} and GC \cite{gated} are better and very close to each other.  SF \cite{structureflow} shows higher performance in SSIM and PSNR scores in large mask sizes. Our approach achieves competitive results compared to the mentioned state-of-the-art methods without using light-weight generators. Table~\ref{celebHQ} reports the quantitative comparison of CelebHQ. Our proposed method outperforms CA, which shows significantly lower performance. Also, it achieves very comparable results to GC and SF that have a larger number of network parameters.

\begin{table}
\centering
\caption{Quantitative evaluation on Places2 dataset with CA \cite{ca}, EC \cite{edge}, DFNet \cite{dfnet}, GC \cite{gated}, SF \cite{structureflow} and our model. (for MAE lower is better, for SSIM and PSNR higher is better). The best scores are indicated in bold.}
\scalebox{0.9}{
\begin{tabular}{|p{0.20cm}|p{1.13cm}|p{0.75cm}|p{0.75cm} |p{0.75cm} |p{0.75cm} |p{0.75cm} |p{0.75cm}|}  
\hline
  & Mask & CA & EC & DFNet & GC & SF & Ours \\ \hline
\hline
\centering
\multirow{4}{*}{\rot {$MAE^-$}} & 10-20\% & 0.019 & 0.013 & 0.010 & 0.011 & 0.012 & \textbf{0.009}\\
 & 20-30\% & 0.033 & 0.022 & 0.019 & 0.018 & 0.019 & \textbf{0.016} \\
 & 30-40\% & 0.048 & 0.031 & 0.028 & 0.026 & 0.026 & \textbf{0.024} \\
 & 40-50\% & 0.075 & 0.053 & 0.045 & 0.045 & 0.044 & \textbf{0.042} \\ \hline
 \hline
\centering
 \multirow{4}{*} {\rot {$SSIM^+$} } & 10-20\% & 0.922  & 0.947 & 0.965 & 0.969 & 0.966 & \textbf{0.971}\\
 & 20-30\% & 0.861 &  0.913 & 0.936 & 0.942 & 0.944 & \textbf{0.946} \\
 & 30-40\% & 0.795 &  0.879 & 0.901 & 0.909 & 0.912 & \textbf{0.916} \\
 & 40-50\% & 0.660 &  0.762 & 0.803 & 0.810 & 0.812 & \textbf{0.816} \\ \hline
 \hline
\centering
 \multirow{4}{*} {  \rot {$PSNR^+$} } & 10-20\% & 26.31 & 27.88 & 29.51 & 30.10 & 30.23 & \textbf{30.62}\\
 & 20-30\% & 23.07 & 25.51 & 26.73 & 27.13 & 27.32 & \textbf{27.71} \\
 & 30-40\% & 20.91 & 23.96 & 24.87 & 25.07 & 25.38 & \textbf{25.74} \\
 & 40-50\% & 18.27 & 20.80 & 21.03 & 21.78 & 21.97 & \textbf{22.55} \\ \hline
\end{tabular}}
\label{places2}
\end{table}

\begin{table}
\centering
\caption{Quantitative evaluation on CelebHQ dataset with CA \cite{ca}, GC \cite{gated}, SF \cite{structureflow} and our model. (for MAE  lower is better, for SSIM and PSNR higher is better). The best scores are indicated in bold.}
\scalebox{1}{
\begin{tabular}{|p{0.5cm}|p{1.2cm}|p{0.8cm} |p{0.8cm} |p{0.8cm}|p{0.8cm} |}  
\hline
  & Mask & CA & GC  &SF & Ours \\ \hline
\hline
\centering
\multirow{4}{*}{\rot {$MAE^-$}} 
 & 10-20\% & 0.014 & 0.009 & 0.011 & \textbf{0.006} \\
 & 20-30\% & 0.024 & 0.014 & 0.015 & \textbf{0.010} \\
 & 30-40\% & 0.033 & 0.020 & 0.018 & \textbf{0.015} \\
 & 40-50\% & 0.052 & 0.031 & 0.028 & \textbf{0.024} \\ \hline
 \hline
\centering
 \multirow{4}{*}{\rot {$SSIM^+$ }} 
 & 10-20\% & 0.950 & 0.982 & 0.984 & \textbf{0.988} \\
 & 20-30\% & 0.918 & 0.968 & 0.971 & \textbf{0.979} \\
 & 30-40\% & 0.881 & 0.940 & 0.950 & \textbf{0.967} \\
 & 40-50\% & 0.796 & 0.905 & 0.912 & \textbf{0.924} \\ \hline
 \hline
\centering
 \multirow{4}{*}{\rot {$PSNR^+$} } 
 & 10-20\% & 28.55 & 32.53 & 33.26 & \textbf{34.64} \\
 & 20-30\% & 25.54 & 29.73 & 30.42 & \textbf{31.79} \\
 & 30-40\% & 23.58 & 27.80 & 28.74 & \textbf{29.81} \\
 & 40-50\% & 21.03 & 25.06 & 25.63 & \textbf{26.64} \\ \hline
\end{tabular}}
\label{celebHQ}
\end{table}

\begin{table}
\centering
\caption{The number of floating points operations, parameters and the runtime on CPU and GPU between CA \cite{ca}, EC \cite{edge}, DFNet \cite{dfnet}, GC \cite{gated} SF \cite{structureflow} and our model on an Intel(R) Core(TM) i7-2600K CPU @ 3.40GHz and an NVidia Titan Xp GPU.}
\scalebox{0.9}{
\begin{tabular}{|c|c|c|c|c|}
\hline
Model & GFLOPS & PARAMS (M) & CPU (ms) & GPU (ms)\\
\hline
\hline
CA & 22.4 & \textbf{2.9M} & 383 & 18 \\
\hline
EC & 122.5 & 21.5M & 704 & 32 \\
\hline
DFNet & 9.7 & 3.3M & 358 & 13 \\
\hline
GC & 39.6 & 4.1M & 490 & 27 \\
\hline
SF & 262.4 & 92.5M & 810 & 36 \\
\hline
Ours & \textbf{9.5} & 3M & \textbf{334} & \textbf{11} \\
\hline
\end{tabular}}
\label{runtime}
\end{table}

\begin{figure}
\centering
\includegraphics[width=0.99\textwidth]{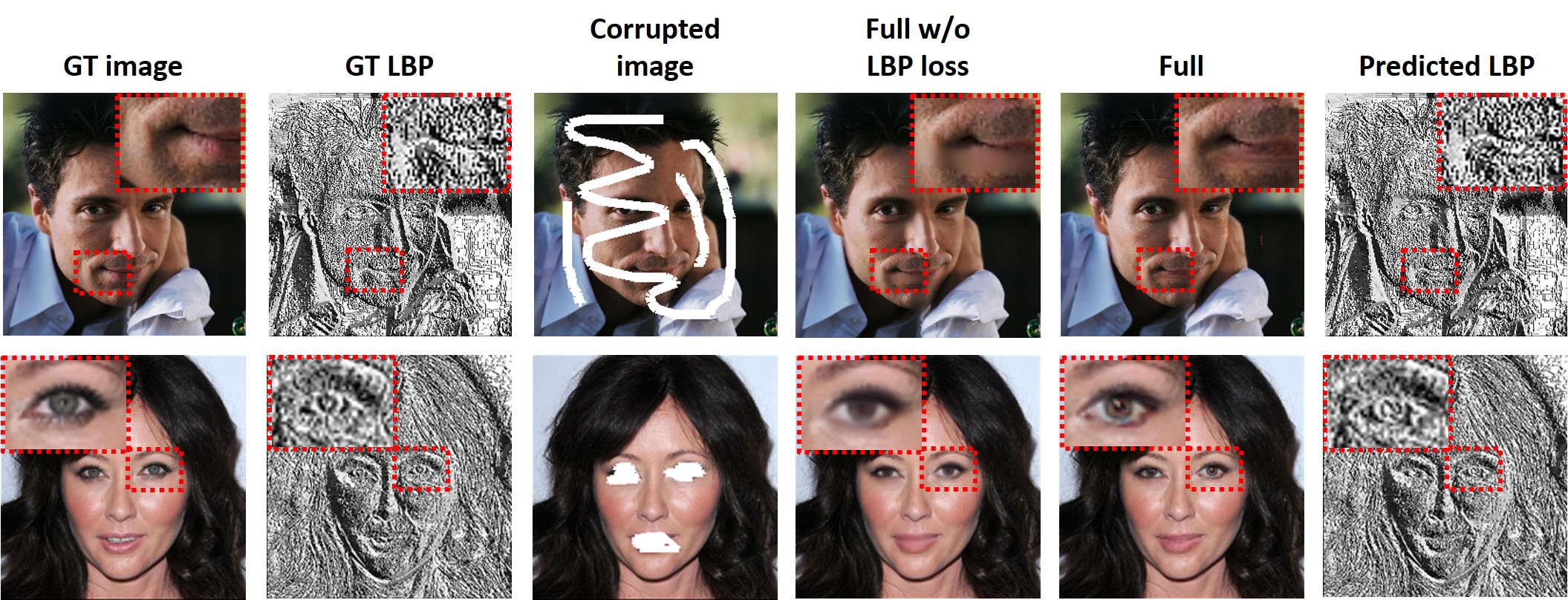}
\caption{Ablation studies showing the effect of the LBP loss on the CelebHQ \cite{style} dataset. From left to right, we show the ground truth image, the LBP image, the input image, the output w/o LBP loss, the output w/ LBP loss, and the predicted LBP image (This figure should be printed in color).}
\label{fig:ablation}
\end{figure}

\begin{figure}
\centering
\includegraphics[width=0.49\textwidth]{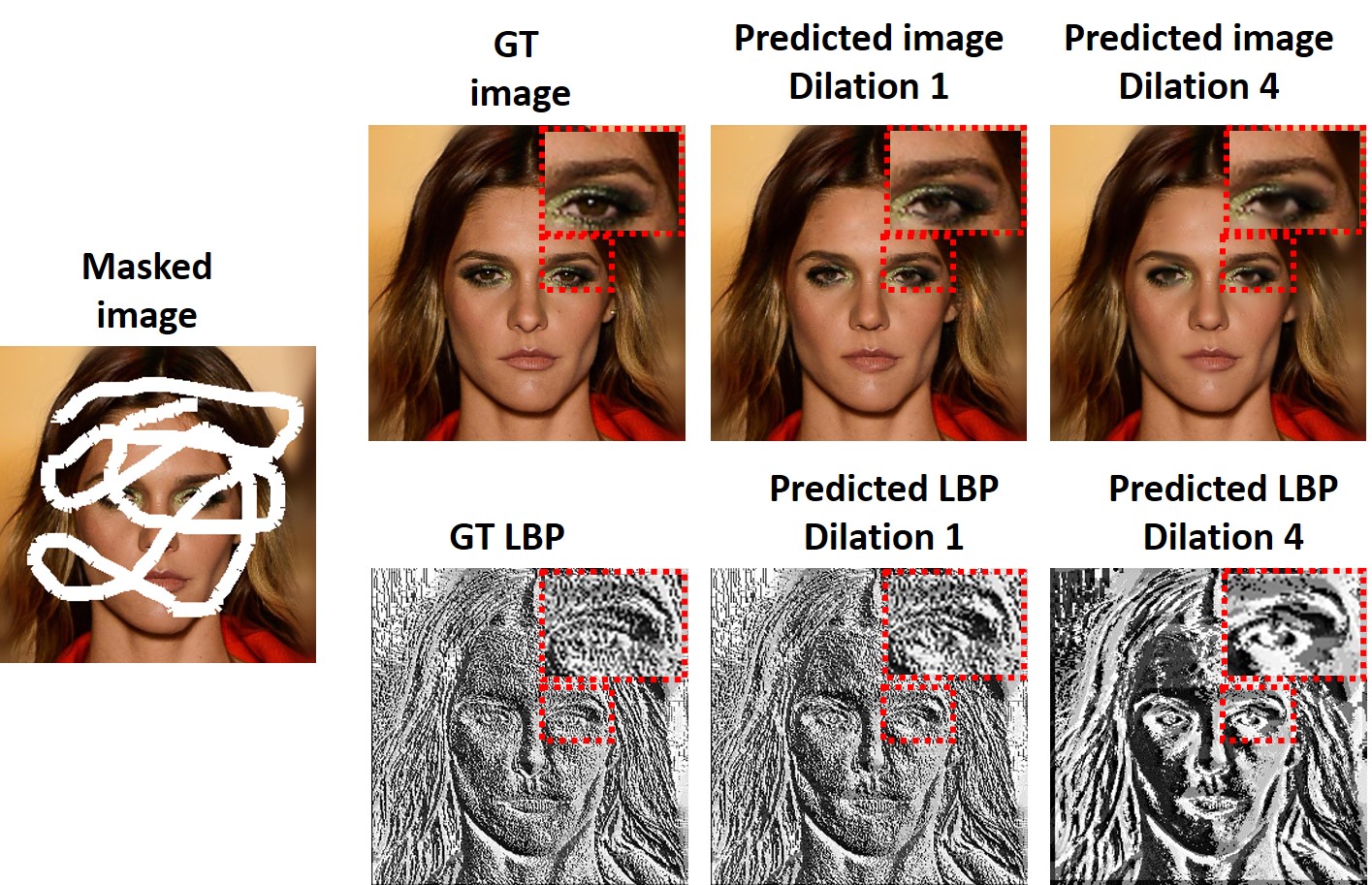}
\caption{Our model final prediction using different dilation factors of the LBP operator (1 and 4) on CelebHQ dataset \cite{style} (This figure should be printed in color).}
\label{fig:lbp_predictions}
\end{figure}

\begin{figure}
\centering
\includegraphics[width=0.99\textwidth]{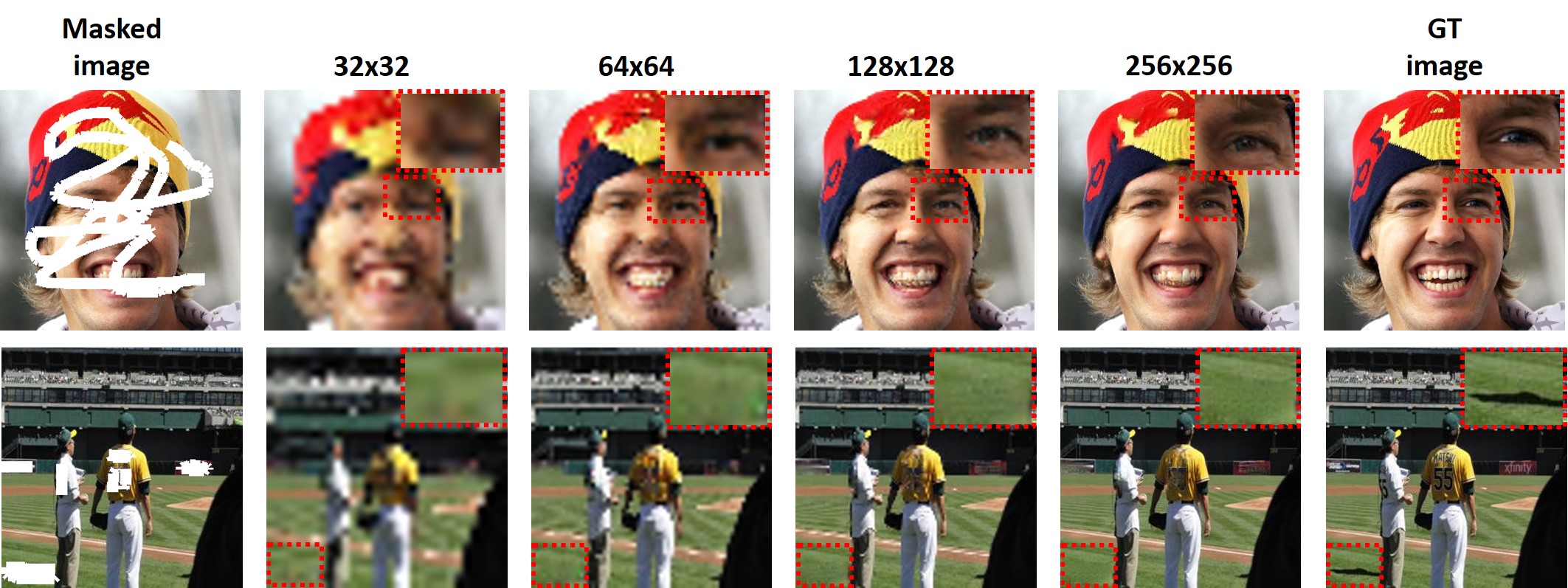}
\caption{Examples from Places2 \cite{places} and CelebHQ \cite{style} datasets of our model intermediate predictions on $32\times 32$, $64\times 64$, $128\times 128$ and $256\times 256$ (This figure should be printed in color).}
\label{fig:intermediate_sizes}
\end{figure}

\subsection{Model efficiency}
\label{parameter}

Table~\ref{runtime} shows the number of floating-point operations in GFLOPS, model parameters in millions and the CPU/GPU runtime in milliseconds. For a fair comparison, we test all the models on the same hardware for 100 iterations to find the mean inference time. We use an Intel(R) Core(TM) i7-2600K CPU @ 3.40GHz and an NVidia Titan Xp GPU. From Table~\ref{places2} and Table~\ref{celebHQ}, we can see that our proposed method performances are very competitive to SF \cite{structureflow} and GC \cite{gated}. But, our model has only 3M parameters and 9.5 (GFLOPS), while SF \cite{structureflow} involves a large number of parameters (92.5M) and  262.4 (GFLOPS) due to the use of two large networks for the smooth and the refined image prediction. Also, GC \cite{gated} has 4.1M parameters and 39.6 (GFLOPS). The result can be explained by that GC \cite{gated} employs costly attention mechanism layers in the refinement network and gated operations (sigmoid activation functions), which augment the number of parameters and GFLOPS, respectively. Our full model is computationally efficient than DFNet \cite{dfnet} that has 9.7 (GFLOPS) and 3.3M parameters. EC \cite{edge} has 21.5M parameters and a computation cost of 122.5 (GFLOPS), which is the result of using two models for edge detection and the refinement network for the final prediction. CA \cite{ca} has the smallest number of parameters (2.9M). However, it has a large computation cost (22.4 GFLOPS) than our model since it involves many attention layers. Besides, it shows the worst performance both in the quantitative and the qualitative comparison. Concerning the inference time, our model shows the best results, highlighting the efficiency of the proposed approach.

\begin{table}
\centering
\caption{Quantitative ablation studies of the LBP loss over the CelebHQ \cite{style} dataset using free-form masks. (for MAE  lower is better, for SSIM and PSNR higher is better). The best scores are indicated in bold.}
\scalebox{1}{
\begin{tabular}{|c|c|c|c|}
\hline
Methods & MAE & SSIM & PSNR \\
\hline
\hline
Full w/o LBP loss & 0.015 & 0.957 & 29.89 \\
\hline
Full w LBP dilation 4 & 0.015 & 0.959 & 30.15 \\
\hline
Full w LBP dilation 1 & \textbf{0.014} & \textbf{0.964} & \textbf{30.72} \\
\hline
Full w perceptual loss  & 0.015 & 0.960 & 30.17 \\
\hline
\end{tabular}}
\label{texture}
\end{table}

\begin{table}
\centering
\caption{Edge prediction metrics over CelebHQ \cite{style} and Places2 \cite{places} datasets in the corrupted regions.}
\scalebox{0.8}{
\begin{tabular}{|c|c|c|c|c|}
\hline
Dataset & Accuracy (\%) & Precision (\%) & Recall (\%) & F1 (\%) \\
\hline
\hline
CelebHQ & 94.9 & 86.2 & 85.2 & 85.7 \\
\hline
Places2 & 93.4 & 84.7 & 84.0 & 84.3 \\
\hline
\end{tabular}}
\label{accuracy}
\end{table}

\begin{figure}
\centering
\includegraphics[width=0.55\textwidth]{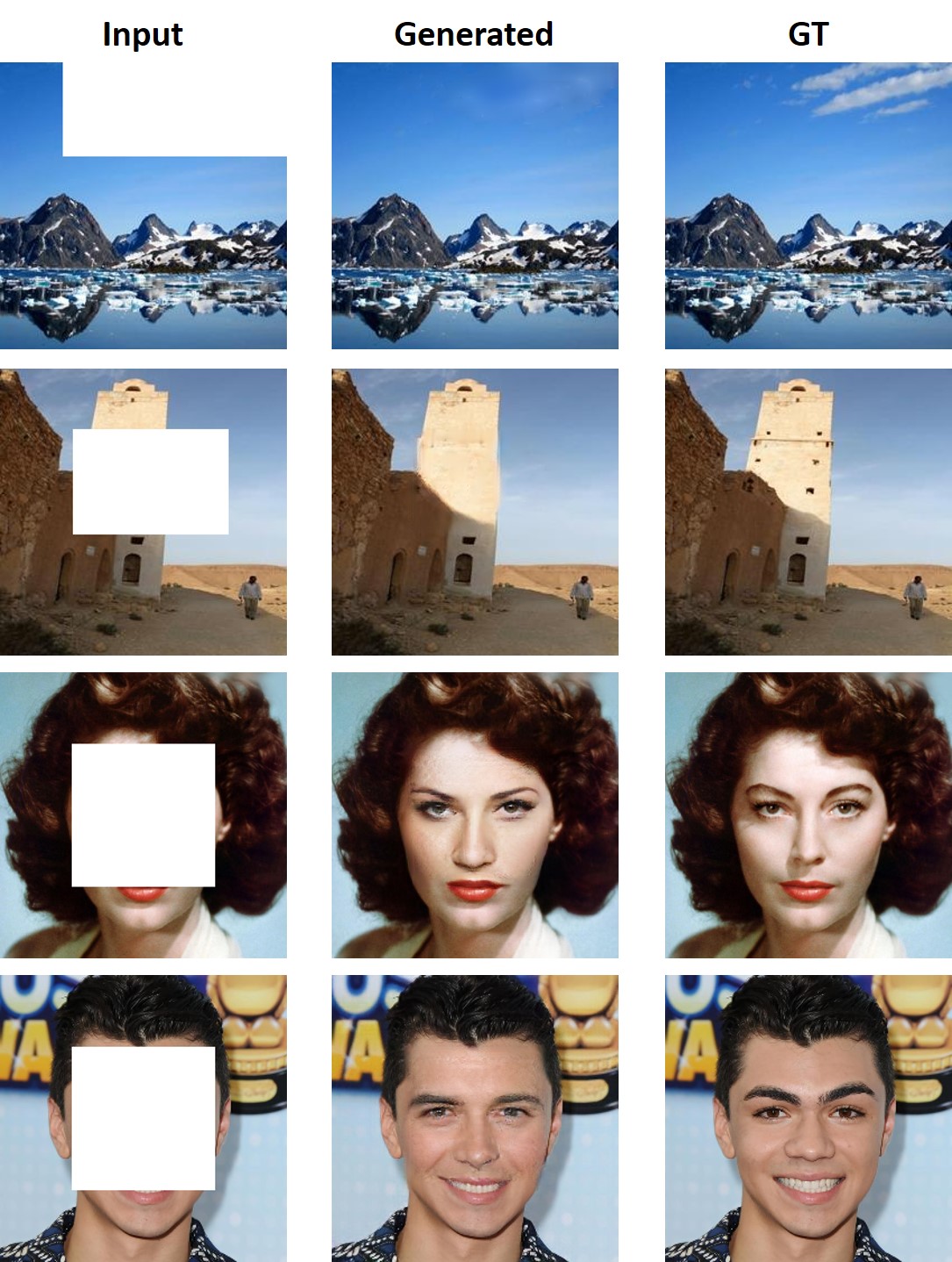}
\caption{Example cases of qualitative results using a rectangular mask on  Places2 \cite{places} and  CelebHQ \cite{style} datasets (This figure should be printed in color).}
\label{fig:blockmask}
\end{figure}

\begin{figure}
\centering
\includegraphics[width=0.55\textwidth]{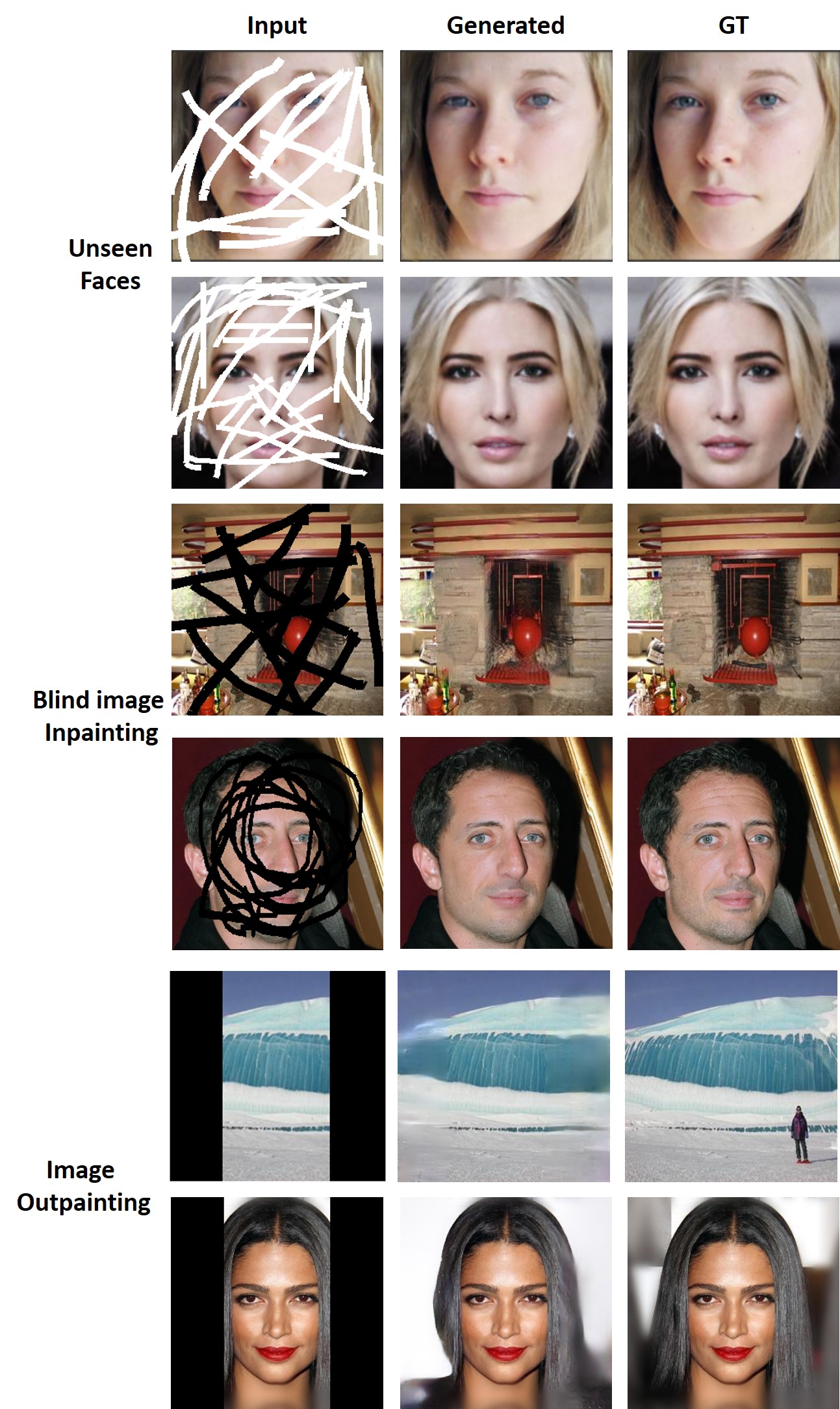}
\caption{Scalability of the proposed approach on different tasks, namely, unseen faces obtained from \cite{c}, blind image inpainting and image outpainting on Places2 \cite{places} and CelebHQ \cite{style} datasets (This figure should be printed in color).}
\label{fig:scalability}
\end{figure}

\begin{figure}
\centering
\includegraphics[width=0.55\textwidth]{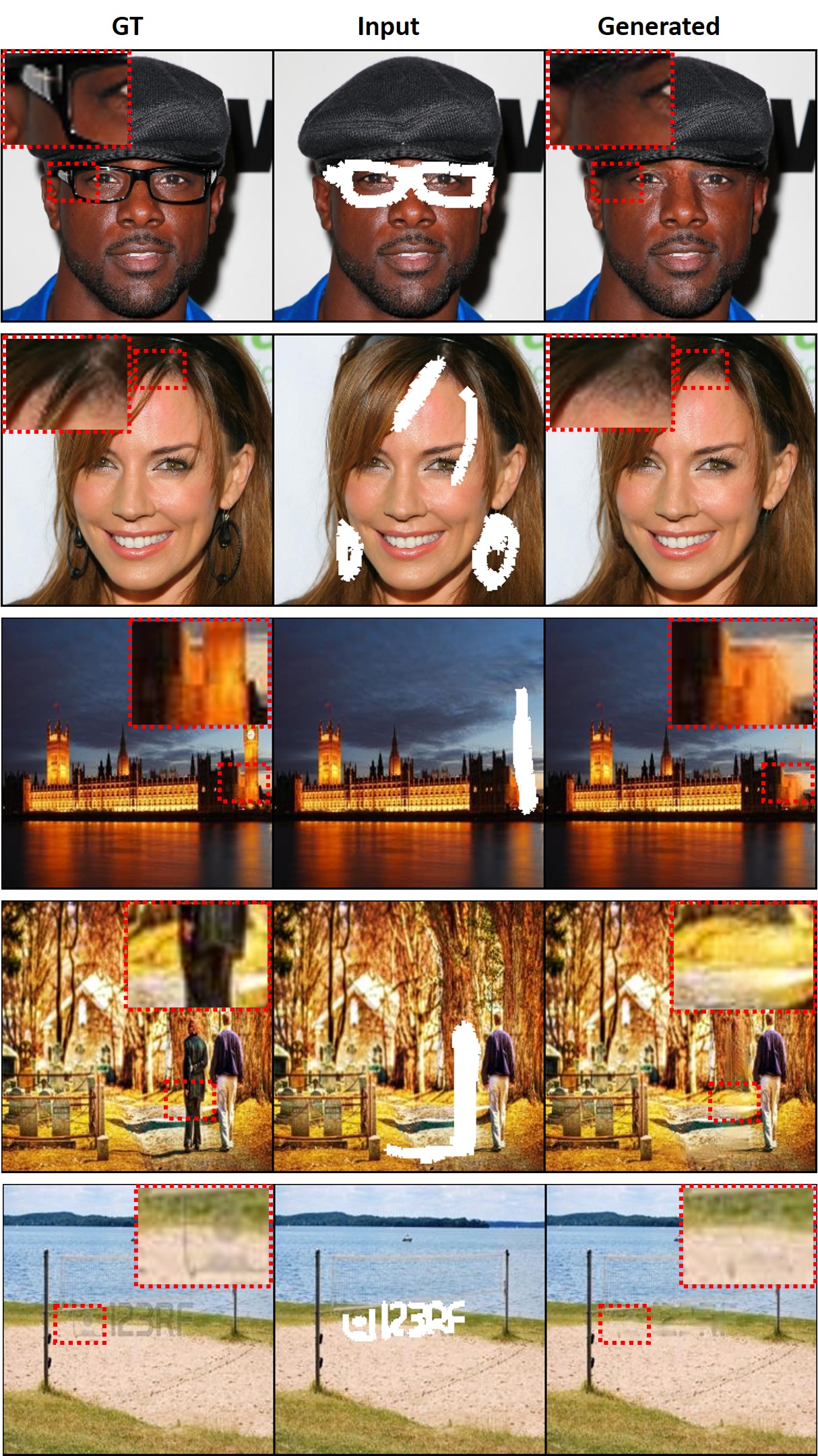}
\caption{Five examples of object removal/editing on CelebHQ \cite{style} and Places2 \cite{places} datasets (This figure should be printed in color).}
\label{fig:usercase}
\end{figure}
\subsection{Ablation study}
\label{ablation}

To further demonstrate the effectiveness of the proposed method and show the contribution of each part to the entire approach, we conduct a set of additional experiments. We investigate the effect of LBP loss function and the LBP operator shape. Also, we compare the proposed LBP-loss function against perceptual loss \cite{perceptual}. We analyze the performances of the four generators, and we evaluate the quality of the generated textures. Finally, we show the scalability of the proposed approach to other applications, including image outpainting and blind image inpainting.

\subsubsection{Effect of the LBP loss}
\label{lbp_loss_effect}
To analyze the contribution of our proposed LBP loss function to the entire approach, we implement two settings of the model, and we show qualitative and quantitative results for each version on the CelebHQ dataset \cite{style}. The first employs only the proposed architecture, while the second adds the LBP loss function to constrain the prediction. We believe that the LBP can describe the texture of the image since the filter comparison operations keep the most meaningful pixels. Table~\ref{texture} indicates that the LBP loss improves the performance and correlates very well with the metrics.  Also, we can see from Fig.~\ref{fig:ablation} that our additional LBP layer restores the image texture and provide realistic images. Note that the images of the first version are plausible and have semantic consistency, which proves the effectiveness of our proposed multi-resolution generators.
 
\subsubsection{Effect of the LBP operator shape}
\label{lbp_shape_effect}
It is well-known that large convolution filters lead to blurriness when applied in the last deconvolution layer. Thus, we fix the filter size to $3\times3$, and we investigate two different LBP dilation factors to show whether they affect the results or not. We can see from Table~\ref{texture} that using a filter with a dilation factor of 1 achieves better results than a filter with dilation of 4 since the latter looks for distant pixels from the desired region, which causes blurriness as seen in Fig.~\ref{fig:lbp_predictions}.

\subsubsection{Perceptual Loss vs. LBP-based loss}
\label{lbp_vs_perceptual}

To further analyze the impact of the LBP-based loss function, we compare the proposed approach to a high-level feature loss baseline. Specifically, we drop the LBP loss and we use the same multi-GANs architecture with the perceptual loss \cite{perceptual}. During training, generated images and real images are fed to a VGG network to produce the intermediate feature maps in different layers. We observe that the perceptual loss drastically increases the training time since it compares high-dimensional feature maps. On the other hand, our approach compares a single LBP feature map, which speeds up the training. Table~\ref{texture} demonstrates the superiority of our full model in all the quantitative metrics.

\subsubsection{Analysis of the four generators}
\label{generators_effect}
Our approach investigates different receptive fields by optimizing the parameters of four progressive generators. In particular, the generator of the higher resolution benefits from the previously inpainted images by the lower ones to learn the global structure of the image. To show the image structure improvement, we analyze the input of the four generators in the two datasets Places2 \cite{places} and CelebHQ \cite{style}. As seen from Fig.~\ref{fig:intermediate_sizes}, as the training advances, the quality of the image is improved, and more meaningful structures (edges and boundaries) appear. Although the images of lower-resolutions are blurry and do not provide sharp texture details, they recover the global structure of the image, which aids the damaged pixels estimation of the next resolution. We can see from Fig.~\ref{fig:intermediate_sizes} that the resolution $32 \times 32$ recovers the global structure of the nose and the eyes. However, the images are still blurry and lack texture details. As the resolution increases, more visually appealing nose and eyes are generated.

\subsubsection{Analysis of the generated texture quality}
\label{texture_quality}
To further evaluate our approach performances, we measure the accuracy of the edge in the corrupted regions for the Places2 \cite{places} and the CelebHQ \cite{style} datasets since edges robustly express the image structure. We use Canny \cite{canny} since it one of the famous edge detectors to find the edges in the generated and the ground truth images. We calculate different metrics on the predicted edge of the corrupted regions to show the percentage of the recovered edges. Table~\ref{accuracy} shows that our approach can restore most of the texture details since it achieves high scores in the precision, recall, accuracy, and F1 measure.

\begin{table}
\centering
\caption{Quantitative evaluation results of the proposed approach on different tasks, including  block-wise masks, blind image inpainting and image outpainting (for MAE lower is better, for SSIM and PSNR higher is better).}
\begin{tabular}{|c|c|c|c|}  
\hline
\textbf{Task}  & \textbf{Metric} & \textbf{Places2} & \textbf{CelebHQ} \\ \hline
\hline

 \multirow{4}{*}{Block-wise masks} & 
$MAE^-$ & 0.043 &  0.025\\
 & $SSIM^+$ &  0.813 & 0.907 \\
 & $PSNR^+$ & 22.16  &  26.31 \\
   \hline

\multirow{4}{*}{Blind inpainting (Free-form mask)} & 
$MAE^-$ &  0.031& 0.013 \\
 & $SSIM^+$ & 0.889 & 0.952 \\
 & $PSNR^+$ &  24.86 & 28.72  \\
 
  \hline
  
\multirow{4}{*} {Outpainting}& 
$MAE^-$ &  0.046 &0.027 \\
 & $SSIM^+$ & 0.802 &  0.886\\
 & $PSNR^+$ & 20.87 &24.66 \\
 \hline

\end{tabular}
\label{table:ablation_scalability}
\end{table}

\subsubsection{Scalability of the proposed method}
\label{scalability}

To confirm the scalability of the proposed method, we conduct four completing experiments. Quantative and qualitative results are shown in Table.~\ref{table:ablation_scalability}, Fig.~\ref{fig:blockmask} and Fig.~\ref{fig:scalability}. 
In the first experiment, we train and test it using block-wise masks. Specifically, a single hole region with a rectangular shape is placed at different locations. Although this experiment is more challenging than the free-form masks, our method shows remarkable relevance between the squared hole and the background on CelebHQ \cite{style}. Also, it still exhibits visually plausible results on both uniform and non-uniform backgrounds on the Places2 dataset \cite{places}.

To verify the generalization capability of the proposed method, we test our pretrained model on images from the internet \cite{c}. Although we run the model on unseen faces, it performs well in generating visually appealing results with realistic textures.

The third experiment evaluates the approach on blind image inpainting. During training, we give only the corrupted image without the mask. We obtain promising results, confirming that the proposed approach can be applied to other real-world applications. Note that the proposed method achieves higher performances in the image inpainting task since the mask guides the model to distinguish between valid and missing pixels. In the last experiment, we investigate the image outpainting task \cite{d}. We mask 1/4 in the left and the right of the image, and we retrain and test our model on CelebHQ \cite{style} and Places2 \cite{places} datasets. This experiment reports lower performances compared to all tasks. We can explain this by the fact that image outpainting includes two challenges: the missing mask channel and having two large separate corrupted regions.
 
\subsection{Interactive editing}
\label{usercase}
Our method allows users to remove unwanted objects by interactively drawing the input masks. At the same time, it can robustly recover the corrupted parts without artifacts. In both cases, the generated images have realistic texture and global semantic consistency. Some results of the interactive inpainting are provided in Fig.~\ref{fig:usercase}. Our approach robustly removes the glasses and face accessories around complex textured objects such as eyes and hair in the CelebHQ \cite{style} dataset. Further, it provides plausible images on the Places2 \cite{places} dataset that includes crowded scenes.

\section{Conclusion}
\label{conclusion}
In this study, we introduce an effective and efficient end-to-end GAN-based framework for image inpainting. Our approach employs progressive efficient generators to stabilize the training and improve the performances. We fill in different image sizes, such that the generators of higher-resolution profit from the previously inpainted regions. Moreover, we demonstrate that the proposed LBP-based loss function constrains image inpainting and enforces texture details. We report quantitative and qualitative comparisons on Places2 and CelebA-HQ datasets. Experimental results that the proposed approach generates realistic images with global structure consistency and fine-grained textures. Also, it outperforms state-of-the-art methods and significantly speeds up the computational time. Furthermore, it shows promising results for other related applications, such as image outpainting and image blind inpainting. In our future work, we are planning to adapt our architecture and loss function to other image-to-image translation tasks, including image super-resolution, image denoising and image deblurring.

\section*{Declaration of Competing Interest}
The authors declare that they have no known competing financial interests or personal relationships that could have appeared to influence the work reported in this paper.

\section*{Acknowledgement}
This research did not receive any specific grant from funding agencies in the public, commercial, or not-for-profit sectors.
\bibliographystyle{cas-model2-names}

\bibliography{cas-refs}

\appendix

\section*{Supplementary material}
\section{Discriminators}
\label{appendix_discriminators}

\begin{table}
\caption{Architecture of the discriminator network}
\begin{tabular}{|c|c|c|c|c|c|}  
\hline
Layer & Dim & Kernel & Stride & Padding & Activation \\ \hline
\hline
\centering
Conv2D & $n$ & $4\times4$ & $2$ & $1$ & LeakyReLU\\
Conv2D & $n \times 2$ & $4\times4$ & $2$ & $1$ & LeakyReLU\\
Conv2D & $n \times 4$ & $4\times4$ & $2$ & $1$ & LeakyReLU\\
Conv2D & $1$ & $4\times4$ & $1$ & $1$ & - \\
 \hline
\end{tabular}
\label{table:discriminator}
\end{table}

Table~\ref{table:discriminator} shows the architecture of the PatchGAN discriminator \cite{pix2pix} where: $n=24$ for the $32\times32$ and the $64\times64$ discriminators, and $n=28$ for the $128\times128$ and the $256\times256$ discriminators. We use a slope of 0.2 in the LeakyReLU activation function. We use Spectral Normalization \cite{spectral} in the convolution layers where: $bias=False$. We initialize the weights using a Gaussian distribution with $gain=0.02$.

\section{Generators}
\label{appendix_generators}

\begin{table}
\caption{Architecture of the $32\times32$  generator network}
\begin{tabular}{|c|c|c|c|c|c|}  
\hline
Layer & Dim & Kernel & Stride & Padding & Activation \\ \hline
\hline
\centering
Conv2D & $24$ & $3\times3$ & $1$ & $1$ & ReLU \\
Conv2D & $48$ & $4\times4$ & $2$ & $1$ & ReLU \\
Conv2D & $48$ & $4\times4$ & $2$ & $1$ & ReLU \\
Conv2D & $96$ & $3\times3$ & $1$ & $1$ & ReLU \\
Conv2D & $96$ & $3\times3$ & $1$ & $1$ & ReLU \\
Conv2D & $96$ & $3\times3$ & $1$ & $1$ & ReLU \\
Conv2D & $96$ & $3\times3$ & $1$ & $1$ & ReLU \\
Conv2D & $96$ & $3\times3$ & $1$ & $1$ & ReLU \\
TConv2D & $48$ & $4\times4$ & $2$ & $1$ & ReLU \\
TConv2D & $24$ & $4\times4$ & $2$ & $1$ & ReLU \\
Conv2D & $3$ & $3\times3$ & $1$ & $1$ & Tanh \\
\hline
\end{tabular}
\label{table:generator32x32}
\end{table}

\begin{table}
\caption{Architecture of the $64\times64$ generator network}
\begin{tabular}{|c|c|c|c|c|c|c|}  
\hline
 Block & Layer & Dim & Kernel & Stride & Padding & Activation \\ \hline
\hline
\multirow{4}{*}{1}
 & Conv2D & $24$ & $3\times3$ & $1$ & $1$ & ReLU \\
 & Conv2D & $48$ & $4\times4$ & $2$ & $1$ & ReLU \\
 & Conv2D & $48$ & $4\times4$ & $2$ & $1$ & ReLU \\
 \hline
 \hline
\centering
\multirow{4}{*}{2}
 & Conv2D & $24$ & $3\times3$ & $1$ & $1$ & ReLU \\
 & Conv2D & $48$ & $4\times4$ & $2$ & $1$ & ReLU \\
 & Conv2D & $48$ & $3\times3$ & $1$ & $1$ & ReLU \\
 \hline
 \hline
\centering
\multirow{4}{*}{3}
 & Conv2D & $96$ & $3\times3$ & $1$ & $1$ & ReLU \\
 & Conv2D & $96$ & $3\times3$ & $1$ & $1$ & ReLU \\
 & Conv2D & $96$ & $3\times3$ & $1$ & $1$ & ReLU \\
 & Conv2D & $96$ & $3\times3$ & $1$ & $1$ & ReLU \\
 & Conv2D & $96$ & $3\times3$ & $1$ & $1$ & ReLU \\
 & TConv2D & $48$ & $4\times4$ & $2$ & $1$ & ReLU \\
 & TConv2D & $24$ & $4\times4$ & $2$ & $1$ & ReLU \\
 & Conv2D & $3$ & $3\times3$ & $1$ & $1$ & Tanh \\
\hline
\end{tabular}
\label{table:generator64x64}
\end{table}

\begin{table}
\caption{Architecture of the $128\times128$ generator network}
\begin{tabular}{|c|c|c|c|c|c|c|}  
\hline
 Block & Layer & Dim & Kernel & Stride & Padding & Activation \\ \hline
\hline
\multirow{4}{*}{1}
 & Conv2D & $28$ & $3\times3$ & $1$ & $1$ & ReLU \\
 & Conv2D & $56$ & $4\times4$ & $2$ & $1$ & ReLU \\
 & Conv2D & $56$ & $4\times4$ & $2$ & $1$ & ReLU \\
 \hline
 \hline
\centering
\multirow{4}{*}{2}
 & Conv2D & $28$ & $3\times3$ & $1$ & $1$ & ReLU \\
 & Conv2D & $56$ & $4\times4$ & $2$ & $1$ & ReLU \\
 & Conv2D & $56$ & $3\times3$ & $1$ & $1$ & ReLU \\
 \hline
 \hline
\centering
\multirow{4}{*}{3}
 & Conv2D & $28$ & $3\times3$ & $1$ & $1$ & ReLU \\
 & Conv2D & $56$ & $3\times3$ & $1$ & $1$ & ReLU \\
 & Conv2D & $56$ & $3\times3$ & $1$ & $1$ & ReLU \\
 \hline
 \hline
\centering
\multirow{4}{*}{4}
 & Conv2D & $112$ & $3\times3$ & $1$ & $1$ & ReLU \\
 & Conv2D & $112$ & $3\times3$ & $1$ & $1$ & ReLU \\
 & Conv2D & $112$ & $3\times3$ & $1$ & $1$ & ReLU \\
 & Conv2D & $112$ & $3\times3$ & $1$ & $1$ & ReLU \\
 & Conv2D & $112$ & $3\times3$ & $1$ & $1$ & ReLU \\
 & TConv2D & $56$ & $4\times4$ & $2$ & $1$ & ReLU \\
 & TConv2D & $28$ & $4\times4$ & $2$ & $1$ & ReLU \\
 & Conv2D & $3$ & $3\times3$ & $1$ & $1$ & Tanh \\
\hline
\end{tabular}
\label{table:generator128x128}
\end{table}

\begin{table}
\caption{Architecture of the $256\times256$ generator network}
\begin{tabular}{|c|c|c|c|c|c|c|}  
\hline
 Block & Layer & Dim & Kernel & Stride & Padding & Activation \\ \hline
\hline
\multirow{4}{*}{1}
 & Conv2D & $28$ & $3\times3$ & $1$ & $1$ & ReLU \\
 & Conv2D & $56$ & $4\times4$ & $2$ & $1$ & ReLU \\
 & Conv2D & $56$ & $4\times4$ & $2$ & $1$ & ReLU \\
 \hline
 \hline
\centering
\multirow{4}{*}{2}
 & Conv2D & $28$ & $3\times3$ & $1$ & $1$ & ReLU \\
 & Conv2D & $56$ & $4\times4$ & $2$ & $1$ & ReLU \\
 & Conv2D & $56$ & $3\times3$ & $1$ & $1$ & ReLU \\
 \hline
 \hline
\centering
\multirow{4}{*}{3}
 & Conv2D & $28$ & $3\times3$ & $1$ & $1$ & ReLU \\
 & Conv2D & $56$ & $3\times3$ & $1$ & $1$ & ReLU \\
 & Conv2D & $56$ & $3\times3$ & $1$ & $1$ & ReLU \\
 \hline
 \hline
\centering
\multirow{4}{*}{4}
 & Conv2D & $28$ & $3\times3$ & $1$ & $1$ & ReLU \\
 & Conv2D & $56$ & $3\times3$ & $1$ & $1$ & ReLU \\
 & Conv2D & $56$ & $3\times3$ & $1$ & $1$ & ReLU \\
 \hline
 \hline
\centering
\multirow{4}{*}{5}
 & Conv2D & $112$ & $3\times3$ & $1$ & $1$ & ReLU \\
 & Conv2D & $112$ & $3\times3$ & $1$ & $1$ & ReLU \\
 & Conv2D & $112$ & $3\times3$ & $1$ & $1$ & ReLU \\
 & Conv2D & $112$ & $3\times3$ & $1$ & $1$ & ReLU \\
 & Conv2D & $112$ & $3\times3$ & $1$ & $1$ & ReLU \\
 & TConv2D & $56$ & $4\times4$ & $2$ & $1$ & ReLU \\
 & TConv2D & $28$ & $4\times4$ & $2$ & $1$ & ReLU \\
 & Conv2D & $3$ & $3\times3$ & $1$ & $1$ & Tanh \\
\hline
\end{tabular}
\label{table:generator256x256}
\end{table}

For all the generators defined in Table~\ref{table:generator32x32}, Table~\ref{table:generator64x64}, Table~\ref{table:generator128x128} and Table~\ref{table:generator256x256}, we use the same weight initialization method used in the discriminator. $TConv2D$ refers to the $ConvTranspose2d$ layer in Pytorch \cite{pytorch}. The $Gray$ function used in Algorithm~\ref{alg:lbp} is done as follows: $Gray(r, g, b) = 0.299 * r + 0.587 * g + 0.110 * b$ where $r, g$, and $b$ are the red, green and blue channels, respectively.

\section{Learning curves}
\label{appendix_curves}
\begin{figure}
\centering
\includegraphics[width=0.7\textwidth]{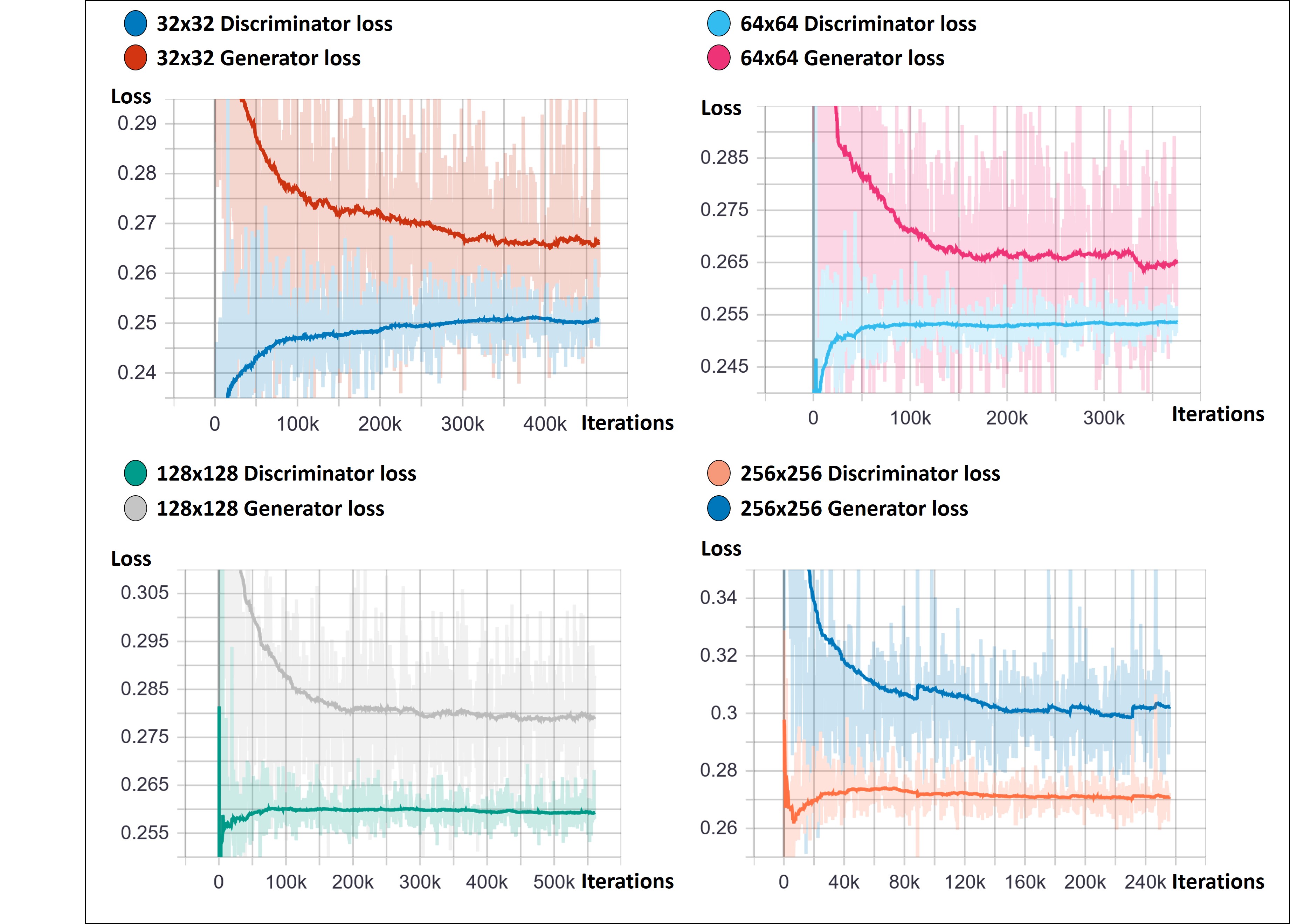}
\caption{The GAN losses of the generators and the discriminators showing a stable training on the four resolutions (This figure should be printed in color).}
\label{fig:loss_gan}
\end{figure}

\begin{figure}
\centering
\includegraphics[width=0.7\textwidth]{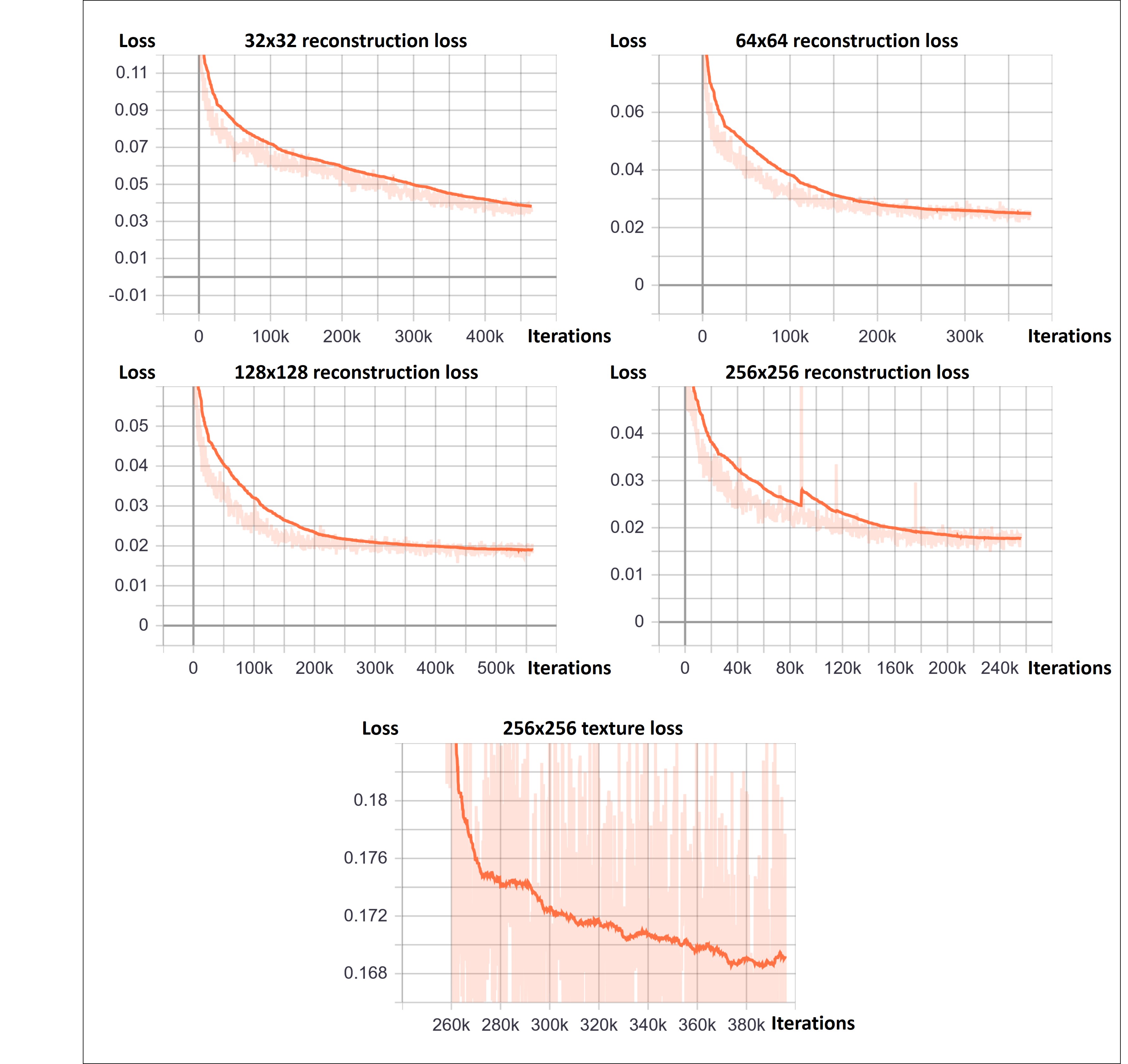}
\caption{The sub-plots represent the image reconstruction loss of the four resolutions and the LBP reconstruction loss of the target resolution ($256\times256$).}
\label{fig:loss_rec}
\end{figure}

We show the training curves of our four generators and discriminators. The successful exploitation of previously inpainted low-resolution images leads to fast training convergence. The loss curves of the adversarial loss show a smooth curve reflecting a stable training and leading to a high visual quality of the generated images. Fig.~\ref{fig:loss_gan} and Fig.~\ref{fig:loss_rec} show the loss values of the generators, the discriminators and the reconstruction loss (image and texture), respectively. During training, we use masks that cover 30-40\% of the image.

\begin{figure}
\centering
\includegraphics[width=0.9 \textwidth]{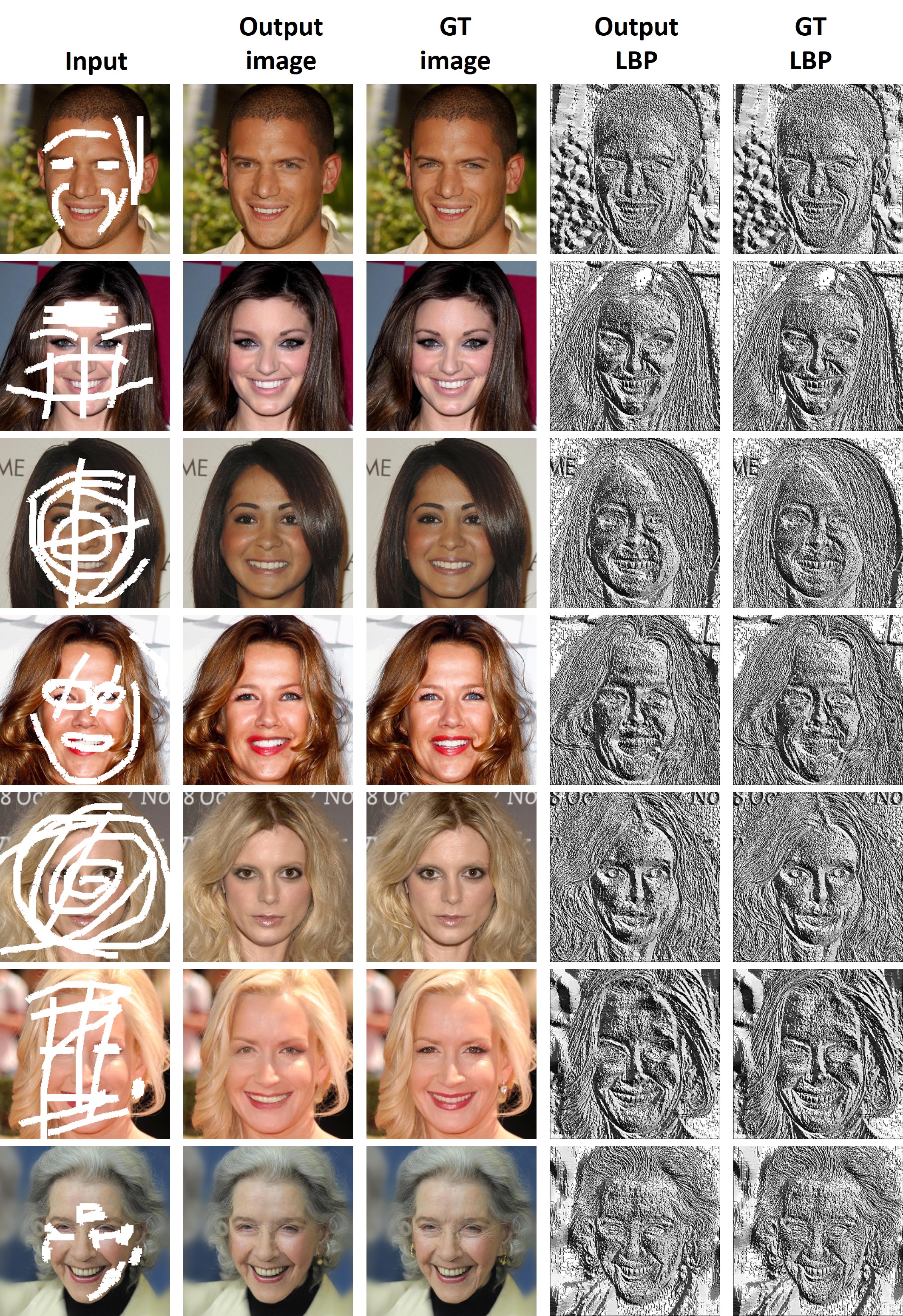}
\caption{Additional results of our model on the CelebHQ \cite{style} dataset (This figure should be printed in color).}
\label{fig:additional_celebhq}
\end{figure}

\begin{figure}
\centering
\includegraphics[width=0.9 \textwidth]{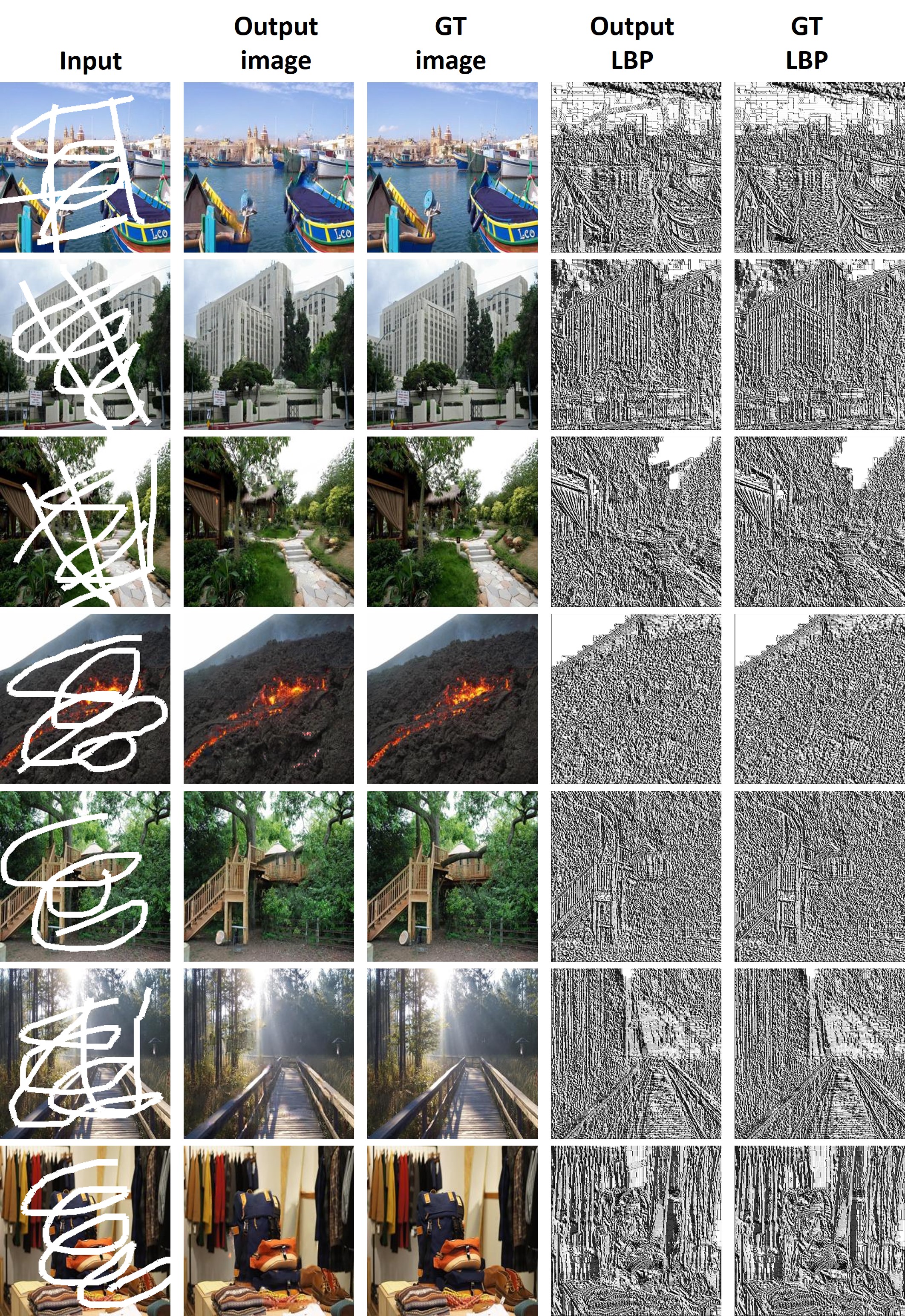}
\caption{Additional results of our model on the Places2 \cite{places} dataset (This figure should be printed in color).}
\label{fig:additional_places2}
\end{figure}

\clearpage



\end{document}